\PassOptionsToPackage{hyphens}{url}
\documentclass[pdfa,a4paper,UKenglish,cleveref,autoref,thm-restate]{tgdk-v2021}

\pdfoutput=1
\nolinenumbers

\hideTGDK
\relatedversiondetails{Published Version}{https://doi.org/10.4230/TGDK.4.2.6}
\hideDateSubmission
\hideDateAcceptance
\DatePublished{2026-09-03 (Journal Version)}

\usepackage{adjustbox}

\title{OM4OV: Leveraging Ontology Matching for Ontology Versioning}
\category{Research}

\author{Zhangcheng Qiang\texorpdfstring{\textsuperscript{*}}{}}{Australian National University, Canberra, Australia}{qzc438@gmail.com}{https://orcid.org/0000-0001-5977-6506}{}
\author{Kerry Taylor}{Australian National University, Canberra, Australia}{kerry.taylor@anu.edu.au}{https://orcid.org/0000-0003-2447-1088}{}
\author{Weiqing Wang}{Monash University, Melbourne, Australia}{teresa.wang@monash.edu}{https://orcid.org/0000-0002-9578-819X}{}

\authorrunning{Z. Qiang, K. Taylor, and W. Wang}
\Copyright{Zhangcheng Qiang, Kerry Taylor, and Weiqing Wang}



\relatedversiondetails{Full Version}{https://arxiv.org/abs/2409.20302}

\supplement{For more information on Supplementary Materials and Additional Resources, including all third-party resources used, we refer to the \hyperref[sec:supp-materials]{list} at the end of this article.}
\supplementdetails[subcategory={OM4OV and Agent-OV}]{Software}{https://github.com/qzc438/ontology-versioning}
\supplementdetails[subcategory={Brick Schema Version Track}]{Software}{https://github.com/qzc438/brick-version-track}

\acknowledgements{The authors thank reviewers for their comments that improved the manuscript. The authors thank Alice Richardson of the Statistical Support Network (Australian National University) for helpful advice on the statistical analysis in this paper. The authors thank Gabe Fierro at Colorado School of Mines for providing useful insights on Brick Schema version changes. The authors also thank the Commonwealth Scientific and Industrial Research Organisation (CSIRO) for supporting this project.}

\additionalMetadata{Generative AI Declaration}{During the preparation of this work, no Generative AI was used to produce the manuscript. However, Generative AI is intrinsic to the research reported here and is properly identified throughout the work.}

\begingroup

\footnotetext{Corresponding author}
\endgroup

\SeriesVolume{4}
\SeriesIssue{2}
\SectionAreaEditor{Stefania Dumbrava, George Fletcher, Matteo Lissandrini, and Riccardo Tommasini}
\SpecialIssue{Data Management for (Knowledge) Graphs}
\SectionAreaNoEds{4}
\ArticleNo{6}

\begin{document}

\maketitle

\begin{abstract}
Due to the dynamics of the Semantic Web, version control is necessary to manage changes in widely used ontologies. Despite the long-standing recognition of ontology versioning (OV) as a crucial component of efficient ontology management, many approaches treat OV as similar to ontology matching (OM) and directly reuse OM systems for OV tasks. In this study, we systematically analyse similarities and differences between OM and OV and formalise an OM4OV framework to offer more advanced OV support. The framework is implemented and evaluated in the state-of-the-art OM system Agent-OM. The experimental results indicate that OM systems can be effectively reused for OV tasks, but without the necessary extensions, can produce skewed measurements, poor performance in detecting update entities, and limited explanations of false mappings. To tackle these issues, we propose an optimisation method called the cross-reference (CR) mechanism, which builds on existing OM alignments to reduce the number of matching candidates and to improve overall OV performance.
\end{abstract}

\newpage

\section{Introduction}
\label{sec: introduction}

Ontologies serve as the backbone of the Semantic Web, providing formal concept descriptions of shared concepts across various applications~\cite{gruber1993translational}. An ontology is not static, and the need for version control arises. While web data is dynamic, any ontology in use must undergo periodic revisions to keep pace with the growth in domain knowledge, modifications to the application adaptation, or corrections to the shared conceptualisation~\cite {klein2001ontology}. For example, it is unrealistic to expect ontologies created in the 1990s to contain concepts such as ``touchscreen'', ``fingerprint sensor'', or ``WiFi antenna''~\cite{haller2020we}. These modifications may cause undesirable deficiencies in artifacts that conform to or that reuse the ontology being changed, leading to severe non-compliance and incompatibility issues in downstream tasks.

Ontology versioning (OV), also known as ontology evolution, aims to distinguish and recognise changes between different ontology versions. This enables data that applies the ontology, ontologies that reuse the ontology, and software that uses the ontology to correctly respond to the ontology version changes~\cite{klein2001ontology}. Although OV plays an important role in ontology management, this area remains underexplored. Many studies~\cite{noy2002promptdiff,noy2004tracking,noy2004ontology,castano2006matchmaking,gross2012computed,zahaf2016alignment,kozierkiewicz2019triggering,kozierkiewicz2019updating,kozierkiewicz2020updating,pietranik2023methods} treat OV as similar to ontology matching (OM) and assume OM systems can be reused for OV tasks. OM is a well-studied problem, so leveraging this deep research legacy for OV should be beneficial. We call this approach ``OM4OV''.

We analyse the complementary relationship between OM and OV tasks and provide a task formulation for the OM4OV framework. We implement the framework in our Agent-OM~\cite{qiang2023agent} system to validate its effectiveness in unifying the OM and OV tasks. We investigate the links between these two tasks in practice and explore optimisations for OM4OV. To the best of our knowledge, this work is the first to systematically analyse the OM4OV framework and to provide theoretical foundations and empirical evidence for its effectiveness and pitfalls. Specifically, our key contributions include:

\begin{itemize}
\item We analyse the similarities and differences between OM and OV. In the OM4OV framework, we describe four disjoint subset alignments produced in the OV process: \textit{remain}, \textit{update}, \textit{add}, and \textit{delete}, each corresponding to a subset of match or non-match in OM. We provide formal definitions and formulations.
\item We extend Agent-OM with the OM4OV framework and call the extended system Agent-OV. We evaluate the performance of OV on the testbed derived from the Ontology Alignment Evaluation Initiative (OAEI)~\cite{oaei} datasets. We find that OM systems can be reused for OV tasks, but extensions are necessary to improve both overall OV performance and its meaningful quality measurement.
\item We propose an optimised OM4OV framework. The new framework introduces a cross-reference mechanism, which builds on prior OM alignments to overcome the pitfalls of the naive OM4OV framework by reducing the number of matching candidates and filtering out ambiguous false mappings, thereby significantly improving overall OV performance.
\end{itemize}

In this paper, we assume the OWL 2 Web Ontology Language~\cite{owl2012} is used to represent an ontology. We use both \emph{concept} and \emph{entity} to refer to OWL classes or properties. We use \emph{subsumption} to mean \texttt{rdfs:subClassOf} or \texttt{rdfs:subPropertyOf} and \emph{equivalence} to mean \texttt{owl:equivalentClass} or \texttt{owl:equivalentProperty} as formal class and property inter-relations.

The remainder of the paper is organised as follows. Section~\ref{sec: related-work} reviews the related work. Section~\ref{sec: formulation} provides the task formulation for the OM4OV framework, while Section~\ref{sec: evaluation} evaluates the OM4OV framework and discusses several common pitfalls for improvement. Section~\ref{sec: optimisation} proposes a cross-reference mechanism as a framework optimisation method. Section~\ref{sec: application} discusses the use of OM4OV in real-world applications. Section~\ref{sec: conclusion} concludes the paper.

\section{Related Work}
\label{sec: related-work}

Version control is recognised as a vital element in ontology management. Different versions need to support interoperability so that version changes do not impede the effective and sustainable use of the ontology.

One manual approach is to extend the ontology itself with internal version information. The Simple HTML Ontology Extensions~\cite{heflin2000dynamic} uses the tag \texttt{BACKWARD-COMPATIBLE-WITH} to record version information. The authors of~\cite{klein2002ontology} argue that a carefully-managed version numbering system embedded in the URI of the ontology (and therefore the fully expanded name of entities defined in the ontology) can minimise the impact of adopting updated versions because unchanged entities will be unaffected in practice. These approaches have been largely adopted by the later OWL languages, where a set of annotation properties related to version information is defined. These include \texttt{owl:versionInfo} and \texttt{owl:priorVersion} to describe the version number of the ontology, \texttt{owl:backwardCompatibleWith} and \texttt{owl:incompatibleWith} to specify an entity's compatible or incompatible corresponding entity in the previous version, and \texttt{owl:DeprecatedClass} and \texttt{owl:DeprecatedProperty} to declare deprecated entities. Later, the ontology language $\tau$OWL~\cite{zekri2016tau} was introduced to extend the OWL triple schema to a 4-tuple schema to represent the versioning of concepts within an ontology. This idea is now incorporated in the new proposals for RDF 1.2 Schema~\cite{rdf12}, which allow time-varying information to be deduced from a temporal dimension within the 4-tuple. An alternative option is to create a separate version log to track changes in versions. Unlike earlier approaches that use an unstructured plain text file, the authors of~\cite{plessers2005ontology} propose a novel approach that uses a version ontology with a change definition language to create a version log. In~\cite{cardoso2020construction}, the authors construct an historical knowledge graph. Storing the version log in a knowledge graph not only avoids repetition, but also enables advanced search functions. The authors of~\cite{sassi2016supporting} argue that version logs may contain redundancy and inconsistent information. They propose a graph-of-relevance approach to interlink different version logs and remove less relevant versions.

Maintaining version information in the ontology can be labour-intensive and error-prone. Either extending the current standard 3-tuple schema or using change logs requires consistent updating over time. In most cases, this process is hand-crafted by the ontology engineer or requires human intervention (e.g. pre-defining a schema or creating a template for the change log). A manual process is more likely to make mistakes and fail to propagate changes to dependent artifacts. Moreover, an ontology may lack or have incomplete version information. In such cases, current approaches have limited capability to detect incorrect or missing version information. These approaches rely on the version information contained in or attached to the ontology by manual methods. If such information is missing or incorrect, automated versioning of ontology concepts becomes necessary.

Reusing OM systems for OV tasks (OM4OV) is a lightweight and fully automatic approach to detecting changes in ontologies. OM4OV is compatible with ontologies that use different recording methods, including URIs or additional versioning triples, as well as ontologies that lack internal version information. The PromptDiff suite~\cite{noy2002promptdiff,noy2004tracking,noy2004ontology} is the canonical work that integrates different heuristic matchers and applies a matching-like comparison for OV. The concept of ``matching'' is later used in the H-Change methodology~\cite{castano2006matchmaking} to assess the assimilation of new concepts into the old ontology. Several studies~\cite{gross2012computed,zahaf2016alignment} note that updates to an ontology will generally create a requirement to update its existing alignments to external ontologies. Methods~\cite{kozierkiewicz2019triggering,kozierkiewicz2019updating,kozierkiewicz2020updating} are proposed to identify changes that affect alignments, and in some cases the framework is proposed to automate alignment updates~\cite{pietranik2023methods}, but these works do not address the discovery of alignments from an ontology to its own updated version.

While existing OM4OV approaches directly use OM systems for OV and focus on developing new algorithms and architectures for ontology version control, they pay less attention to understanding the underlying differences between OM and OV. OM is not exactly the same as OV. OM determines \textit{matched} entities between two different ontologies, while OV can distinguish \textit{add}, \textit{delete}, \textit{remain}, and \textit{update} changes over different versions of one ontology. The OM input consists of two distinct ontologies, whereas the OV input is expected to consist of two versions of a single ontology. The output of OM is a set of entity mappings, whereas the output of OV comprises four sets, each defined by the nature of the change made. These differences may affect the adaptability of reusing OM systems for OV tasks.

\section{Task Formulation}
\label{sec: formulation}

We now draw on concepts in OM to formalise OV tasks. Given an ontology $O$, we say that some entity $e\in O$ when $e$ is of interest to ontology matching or ontology versioning, typically representing a concept of the domain of interest being modelled by $O$. We consider an entity to be the full URI of the reference in an ontology, or its equivalent abbreviation via a prefix, for example \texttt{cmt:ProgramCommitteeChair}. We consider an \textit{entity name} to be the substring of an entity that provides a meaningful description of the entity (e.g. \textit{ProgramCommitteeChair}), excluding any entity prefix or expanded URI (e.g. \texttt{cmt:ProgramCommitteeChair} is not an entity name).

\subsection{OM Preliminaries}

OM identifies correspondences between entities of two ontologies~\cite{euzenat2007ontology}. Formally, given a source ontology ($O_s$) and a target ontology ($O_t$), the OM task is to find an alignment $A$ (i.e. a set of mappings) with respect to a given similarity threshold $s \in [0,1]$, defined as~\cite{euzenat2007semantic}: $A = \{(e_1, e_2, r, c) \mid e_1 \in O_s, e_2 \in O_t, s \leq c \leq 1\}$, where $e_1$ and $e_2$ are ontology entities in $O_s$ and $O_t$ respectively, $r$ is the relation between $e_1$ and $e_2$ that may be equivalence ($\equiv$) or subsumption ($\subseteq$), and $c$ is the confidence. For example, \texttt{(cmt:ProgramCommitteeChair, confof:Chair\_PC, $\equiv$, 0.90)} means \texttt{cmt:ProgramCommitteeChair} and \texttt{confof:Chair\_PC} are equivalent with a confidence of 0.90.

An entity match is asserted by a matching system according to its own internal algorithm, its own interpretation of ``match'', and its own assessment of confidence. In general for matching, a source entity may occur in multiple mappings that vary with respect to the target entity, the relation, and the confidence. In this study, we consider only alignments made by matching systems that are comprised of one-to-one equivalence mappings, admitting only matches with the highest $c$ as determined by the matching system and requiring the matching system to resolve tied confidence scores. Typically, equivalence relations are determined by lexical or embedding similarity in matching systems, although logical entailment is equally admissible in our study. OM performance is measured by comparing the matching system alignment (A) with a reference alignment (R) using precision, recall, and F1 score (see Section~\ref{sub-sec: evaluation-criteria} for details).

\subsection{OM4OV Formalisation}

\begin{figure}[htbp]
\includegraphics[width=1\linewidth]{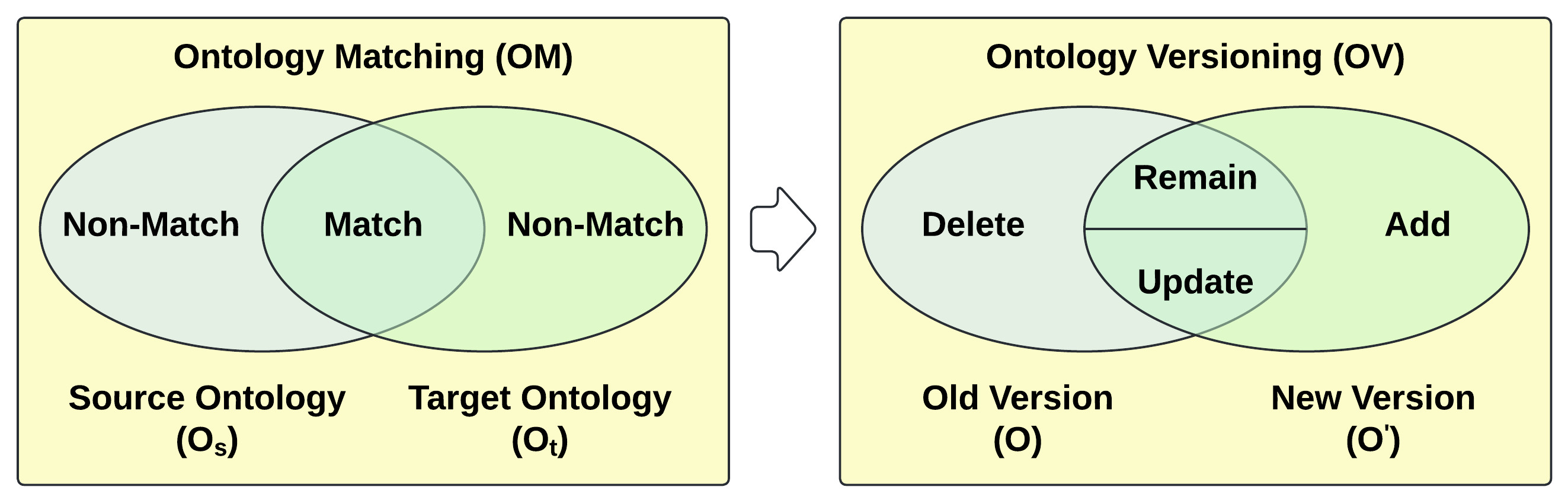}
\caption{Overview of the OM4OV framework. The sets are comprised of entities, with the matched intersection containing entities that are considered equivalent by an ontology alignment.}
\label{fig: om4ov}
\end{figure}

OV identifies correspondences between two versions of a single ontology~\cite{klein2001ontology}. Given an old version of an ontology ($O$) and a new version of the same ontology ($O'$), the OV task can be considered as finding an alignment (A) from the old ontology to the new one. As illustrated in Figure~\ref{fig: om4ov}, while an alignment for OM only mentions matched entities, OV needs to recognise both matched and non-matched entities. Furthermore, OV matched entities are comprised of two subsets: \textit{remain} and \textit{update}, while non-matched entities are comprised of \textit{add} and \textit{delete} entities. While some of these sets are comprised of entities, and others of mappings, when convenient henceforward, we will loosely refer to all of these as either alignments or sets of entities interchangeably.

An OV task can be formalised as finding an alignment and a non-alignment, $A_{match}$ and $A_{non-match}$ respectively, between different ontology versions $O$ and $O'$ with respect to a given similarity threshold $s \in [0,1]$. OV focuses only on the equivalence relation; subsumption mappings are beyond the scope of versioning.
\begin{definition}
Given ontologies $O$ and $O'$ and similarity threshold $s \in [0,1]$, the OV task is to find  $A_{match}$ and $A_{non-match}$ as follows:
\begin{equation}
\begin{aligned}
A_{match} = \{(e_1, e_2, c) &\mid e_1 \in O, e_2 \in O', s \leq c \leq 1 \\ &\mbox { and there is no }  (e_1, e_i, c') \mbox{ with } e_i \neq  e_2 \mbox { or with }  c \neq c'  \\ &\mbox{ and there is no } (e_j, e_2, c') \mbox{ with } e_j \neq  e_1 \mbox { or with }  c \neq c' \}  \\
A_{non-match} = \{e \mid e \in O &\cup O' \mbox{ and there is no }  (e_1,e_2,c) \in A_{match} \mbox{ with } e \in \{e_1,e_2\}\}  \\
\end{aligned}
\end{equation}
We say that $e1$ and $e2$ are equivalent whenever there is a $c$ satisfying $(e1,e2,c) \in A_{match}$.
\end{definition}
For convenience, we will sometimes refer to the elements of $A_{match}$ as \emph{entities} rather than \emph{mappings}, by which we mean entities that are mentioned in mappings in $A_{match}$.

The \textit{update} and \textit{remain} entities are entities successfully matched over different ontology versions. The \textit{update} entities are, in general, the closest near-equivalent matches of entities from different ontology versions. We recommend that matching systems assign lower confidence to entity matches when either (a) structural changes have been made near the entity, such as changes to class membership or property restrictions, or (b) the entity's name has changed.

We recommend that all other successful matches not satisfying either (a) or (b) are assigned very high confidence by matching systems, most especially when a so-called \emph{trivial} match is found. These matches are then interpreted as \textit{remain} entities for OV. In particular, we say that an entity is interpreted as not changed in OV for any successful match where the matching system reports a confidence of 1. Further, we recognise that some OV systems may prefer, additionally, to interpret an entity as not changed when the confidence is only slightly less than 1.
\begin{definition}
$A_{remain} (A\odot)$ and $A_{update} (A\otimes)$ are defined as:\footnotemark
\footnotetext{In some cases, OV systems may consider matches that are near-enough with a confidence slightly less than 1 to be perfect matches. Without loss of generality, we assume in such cases that the OV system reassigns $c$ to be 1.}
\begin{equation}
\begin{aligned}
&A\odot = \{(e_1, e_2, c) \mid (e_1, e_2, c) \in A_{match} \mbox{ and } c = 1\}     \\
&A\otimes = \{(e_1, e_2, c) \mid (e_1, e_2, c)  \in A_{match} \mbox{ and } c < 1\}  \\
\end{aligned}
\end{equation}
\end{definition}

We can see that $A_{match} = A\odot \cup A\otimes$. By definition, $A\odot$ and $A\otimes$ are disjoint.

The \textit{add} and \textit{delete} entities are actually non-matched entities between different ontology versions. The \textit{add} entities are the entities in $O'$ that have no matches in $O$. Similarly, we can interpret the \textit{delete} entities as entities in $O$ that have no matches in $O'$.
\begin{definition}
$A_{add} (A\oplus)$ and $A_{delete} (A\ominus)$ are defined as:
\begin{equation}
\begin{aligned}
&A\oplus = \{e \mid e \in O'\mbox{ and } e \in A_{non-match} \}      \\
&A\ominus = \{e \mid e \in O\mbox{ and } e \in A_{non-match} \}      \\
\end{aligned}
\end{equation}
\end{definition}
We can see that $A_{non-match} = A\oplus \cup A\ominus$. By definition, $A\oplus$ and $A\ominus$ are disjoint.

\section{OM4OV Evaluation}
\label{sec: evaluation}

We now consider the evaluation of OM4OV. In the literature, OV are evaluated using OM measures, which we refine here to give better insights into OV performance.

\subsection{Dataset Construction}
\label{sub-sec: dataset-construction}

\begin{figure}[htbp]
\centering
\includegraphics[width=1\linewidth]{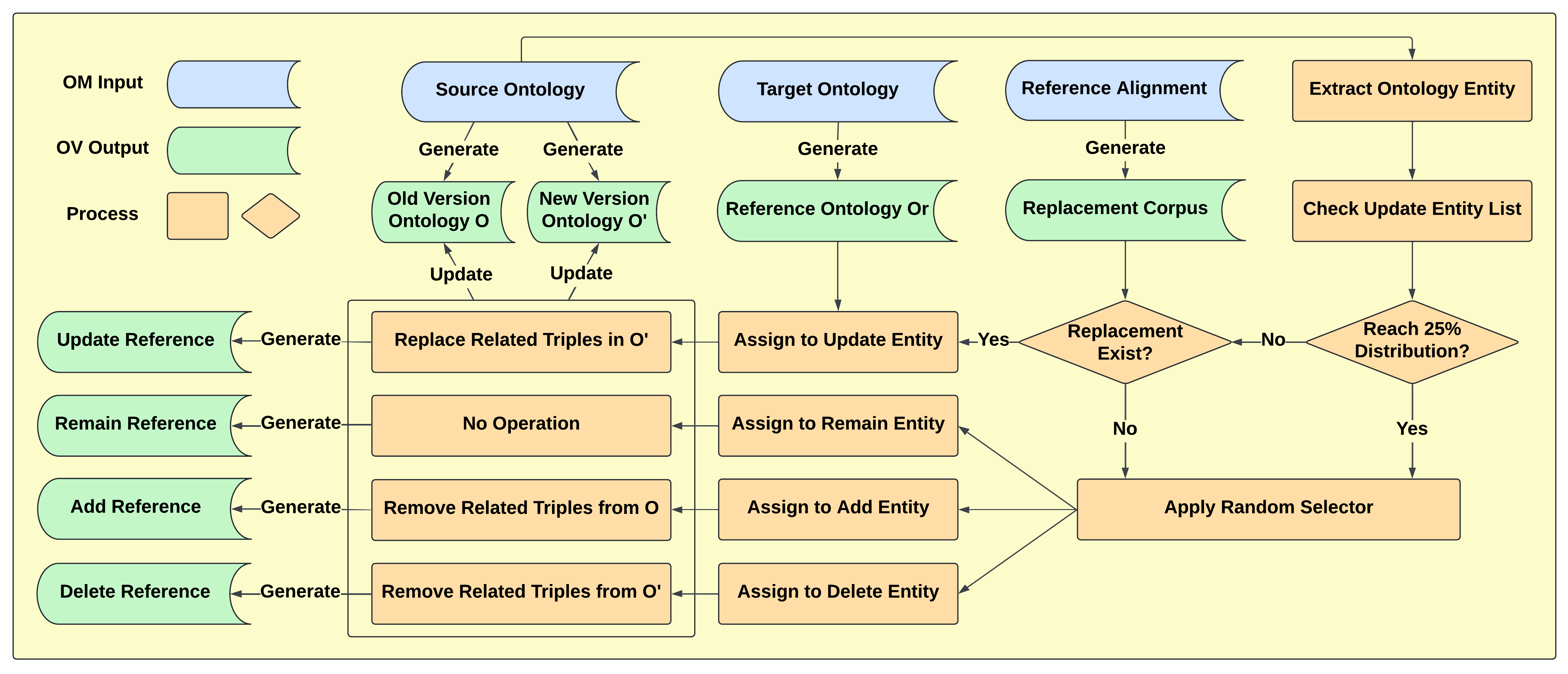}
\caption{Generation of OM4OV datasets from initial OM inputs of a source ontology, a target ontology, and a reference alignment, and outputs of versioning ontologies and reference alignments with four categories. Optionally, a reference ontology and a replacement corpus may be produced.}
\label{fig: dataset-construction}
\end{figure}

There is a dearth of benchmark datasets for evaluating OV. The Ontology Alignment Evaluation Initiative (OAEI) contains several datasets related to OM tasks, but none are specifically designed for OV tasks. We propose an approach to constructing synthetic OV datasets from real-world OM datasets. Figure~\ref{fig: dataset-construction} illustrates the generation of OM4OV datasets.

\begin{bracketenumerate}
\item The original OAEI datasets for OM provide two ontologies, the \textit{source} $O_s$ and the \textit{target} $O_t$. We choose either one as the ontology for versioning and duplicate it into $O$ and $O'$. Another ontology will be used later to provide references for synthetic \textit{update} entities. We duplicate this ontology into $O_r$.
\item There are four possible entity changes in OV tasks: \textit{remain}, \textit{update}, \textit{add}, and \textit{delete}. We retrieve all ontology entities from $O$ or $O'$ (since they are identical at this stage). Each ontology entity will be randomly assigned to one of these categories, and the four sets are pairwise disjoint by construction. In real-world OV, entity updates are relatively infrequent. We allow a user-defined parameter to fix the proportion of update entities in the synthetic dataset, defaulting to 25\%. The update entities are randomly selected from the replacement corpus, so the number of update entities is limited by the size of the replacement corpus, which may be less than the user-defined parameter value. For this reason, selection of update entities is prioritised over other categories, and excess entities are redistributed evenly over the other three categories.
\item For each ontology entity $e$ assigned to a category, the corresponding dataset operations are described as follows. While there is no operation for \textit{remain} entities, entities assigned to \textit{add} and \textit{delete} are deleted from all triples that mention them in $O$ or $O'$ respectively. Each \textit{update} mapping requires that all occurrences of its domain entity in triples in $O'$ are replaced by the corresponding co-domain entity of the mapping. We generate four corresponding versioning references based on the entity assignments. These files provide the ground truth for evaluation.
\end{bracketenumerate}

\noindent To ensure the replaced entity is reasonable and reflects real-world facts, we use equivalent classes or properties specified in the reference alignment. For example, we could replace \texttt{cmt:ProgramCommitteeChair} with its equivalent class \texttt{confof:Chair\_PC}. This operation is cascaded, meaning that we update not only the entity name but also its related triples, such as changing the semantic information from \texttt{(cmt:ProgramCommitteeChair, rdfs:subClassOf, cmt:ProgramCommitteeMember)} to \texttt{(confof:Chair\_PC, rdfs:subClassOf, confof:Person)}.\linebreak  While this procedure can be unlimited, we restrict it to extract only the triples directly related to the entity (i.e. a 1-hop subgraph) in the ontology $O_r$. For entities whose names are unique identifiers or codes (and not textually meaningful names), we retrieve their meaningful descriptions from annotation properties. For example, we extract an additional annotation triple \texttt{(mouse:MA\_0000270, rdfs:label, ``eyelid tarsus'')} for the entity \texttt{mouse:MA\_0000270}. After extraction, we also need to update the prefix from  \textit{``confof''} to \textit{``cmt''} where the new entity URI will be \texttt{cmt:Chair\_PC} and the replacement triple will be \texttt{(cmt:Chair\_PC, rdfs:subClassOf, cmt:Person)}. Finally, we add these new triples to the existing ontology $O'$. Note that where entity names do not change but structural changes have been made near the entity, this will be represented as an entity update. Table~\ref{tab: dataset-update} summarises the steps to construct an \textit{update} entity $e$ in the benchmark dataset.

\begin{table}[htbp]
\centering
\caption{Steps to construct an \textit{update} entity $e$ in the benchmark dataset.}
\label{tab: dataset-update}
\begin{tabular}{|c|l|}
\hline
\multirow{1}{*}{\textbf{Step}}  & \multirow{1}{*}{\textbf{Dataset Operation}}                               \\ \hline
\multicolumn{1}{|c|}{\#1}       & Remove all the triples related to $e$ in $O'$.                            \\ \hline
\multicolumn{1}{|c|}{\#2}       & Find the equivalent class/property $e_r$ in \textit{replacement corpus}.  \\ \hline
\multirow{2}{*}{\#3}            & Extract all the triples related to $e_r$ in $O_r$.                        \\ \cline{2-2}
& (Optional) Extract annotation triples related to $e_r$ in $O_r$.                                          \\ \hline
\multicolumn{1}{|c|}{\#4}       & Update the prefix from $e_r$ to $e$ and create a new entity $e'$.         \\ \hline
\multicolumn{1}{|c|}{\#5}       & Add all the triples related to $e'$ to $O'$.                              \\ \hline
\end{tabular}
\end{table}

Unlike the original OAEI datasets used for OM, we introduce randomness to ensure that the synthetic OAEI datasets for OV are different each time they are constructed. This suits the dynamic nature of OV, where the changes vary between different versions. For this reason, we consider the new OAEI datasets for OV more like a testbed, as they can simulate a variety of situations for OV tasks. When reproducibility is required, we recommend using a fixed seed to control the randomness.

Table~\ref{tab: ov-testbed} details the OAEI tracks selected for the OV testbed. The current version of the OV testbed contains a total of 11 distinct ontologies from three different OAEI tracks. The anatomy track contains two large ontologies, while the MSE track has two medium ontologies, and the conference track has 7 small ontologies. We exclude EMMO~\cite{emmo} from the MSE track because it assigns opaque symbolic codes to entity names, using a naming convention that differs from MaterialInformation~\cite{ashino2010materials} and MatOnto~\cite{matonto}, which use meaningful entity names. Figure~\ref{fig: ov-testbed} shows the number of entities in each ontology in each track.

\noindent
\begin{minipage}[t]{0.64\textwidth}
\centering
\captionof{table}{Selected OAEI tracks for the OV testbed.}
\label{tab: ov-testbed}
\adjustbox{width=\textwidth, valign=t}{
\renewcommand{\arraystretch}{1.3}
\begin{tabular}{|c|c|c|}
\hline
\multirow{1}{*}{\textbf{Track}}     & \multirow{1}{*}{\textbf{Domain}}  & \multirow{1}{*}{\textbf{\# of Ontologies}}    \\ \hline
\multicolumn{1}{|c|}{Anatomy}       & Human and Mouse Anatomy           & 2                                             \\ \hline
\multicolumn{1}{|c|}{Conference}    & Research Conference               & 7                                             \\ \hline
\multicolumn{1}{|c|}{MSE}           & Materials Science and Engineering & 2                                             \\ \hline
\end{tabular}
}
\end{minipage}
\begin{minipage}[t]{0.35\textwidth}
\centering
\captionof{figure}{Number of entities.}
\label{fig: ov-testbed}
\adjustbox{valign=t}{
\includegraphics[width=1\linewidth]{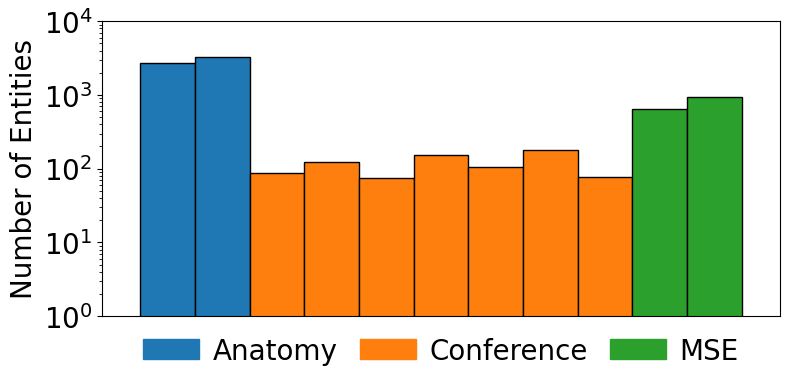}
}
\end{minipage}

\subsection{Evaluation System}
\label{sub-sec: evaluation-system}

We choose the matching system Agent-OM and its extended version Agent-OV to demonstrate the OM4OV framework. Agent-OM is one of the flagship OM systems powered by large language models (LLMs) and multi-agent architectures. Its foundation framework is designed for traditional OM tasks. We extend Agent-OM with the OM4OV framework so that it can be used for OV tasks. Inherited from Agent-OM, Agent-OV supports a wide range of LLMs, including commercial API-accessed LLMs OpenAI GPT~\cite{llm-gpt} and Anthropic Claude~\cite{llm-claude}, as well as open-source LLMs Meta Llama~\cite{llm-llama}, Alibaba Qwen~\cite{llm-qwen}, Google Gemma~\cite{llm-gemma}, and ChatGLM~\cite{glm2024chatglm}. Inherited from  Agent-OM, $similarity\_threshold$ and $top@k$ are the adjustable key hyperparameters for Agent-OV.

As an aside, we observe that confidence scores for successful matches as determined by OM systems may be unhelpful for direct use in OV for distinguishing \textit{remain} entities from \textit{update} entities. This is because, in practice,  the confidence boundary to best distinguish  \textit{remain} and \textit{update} entities varies in different domains and datasets. In Agent-OV, we handle this problem in the following way. A matched entity is determined to be a \textit{remain} entity when its syntactic name and lexical descriptions are unchanged. Otherwise it is determined to be an \textit{update} entity.  We observe that this approach means that Agent-OV usually considers entities to be \textit{remain} entities when they have structural changes but conserve syntactic names and lexical descriptions. This approach will tend to consider ambiguous entities as \textit{remain} in preference to \textit{update}, so tending to trade off performance in \textit{remain} versus performance in \textit{update}. While other OV systems may handle this differently, we propose that all OM4OV systems could benefit from this approach to partitioning successful matches into \textit{remain} entities and \textit{update} entities.

\subsection{Evaluation Criteria}
\label{sub-sec: evaluation-criteria}

OM typically measures performance using precision, recall, and the F1 score. Given a gold standard reference (R) and a system-discovered alignment (A), precision measures matching correctness and recall measures matching completeness, while F1 score offers a harmonic mean to balance correctness and completeness. OV can reuse these measures, but they need to be extended into four sub-measures for add, delete, remain, and update performance. While an OV method will trade off performance over each of these sets, an OV application problem may well prioritise accuracy on some over others.
\begin{equation}
\begin{aligned}
Precision = \frac{|A \cap R|}{|A|} \quad Recall = \frac{|A \cap R|}{|R|} \quad F_1 \ Score= \frac{2}{Precision^{-1} + Recall^{-1}}
\end{aligned}
\end{equation}

\subsection{Results}
\label{sub-sec: results}

We choose to use the Meta open-source model llama-3-8b~\cite{llama-3-8b} for our experiments in this paper, eschewing the higher-end models that offer small improvements that may not justify the financial investment. We use embeddings from the same LLM model to further avoid expensive API calls and thereby provide an entirely free-to-use version of Agent-OV. The hyperparameter settings are $similarity\_threshold = 0.90$ and $top@k = 3$ across all the system alignments generated. We set $top@k = 3$ based on empirical experience that 3-5 works well for Agent-OM. We experiment with the alternative values of the similarity threshold later in this paper. The fixed seed for the OV testbed is set to 42. Due to the non-determinism of LLMs, repeated experiments can produce insignificantly different results.

\begin{figure}[!t]
\centering
\includegraphics[width=0.325\textwidth]{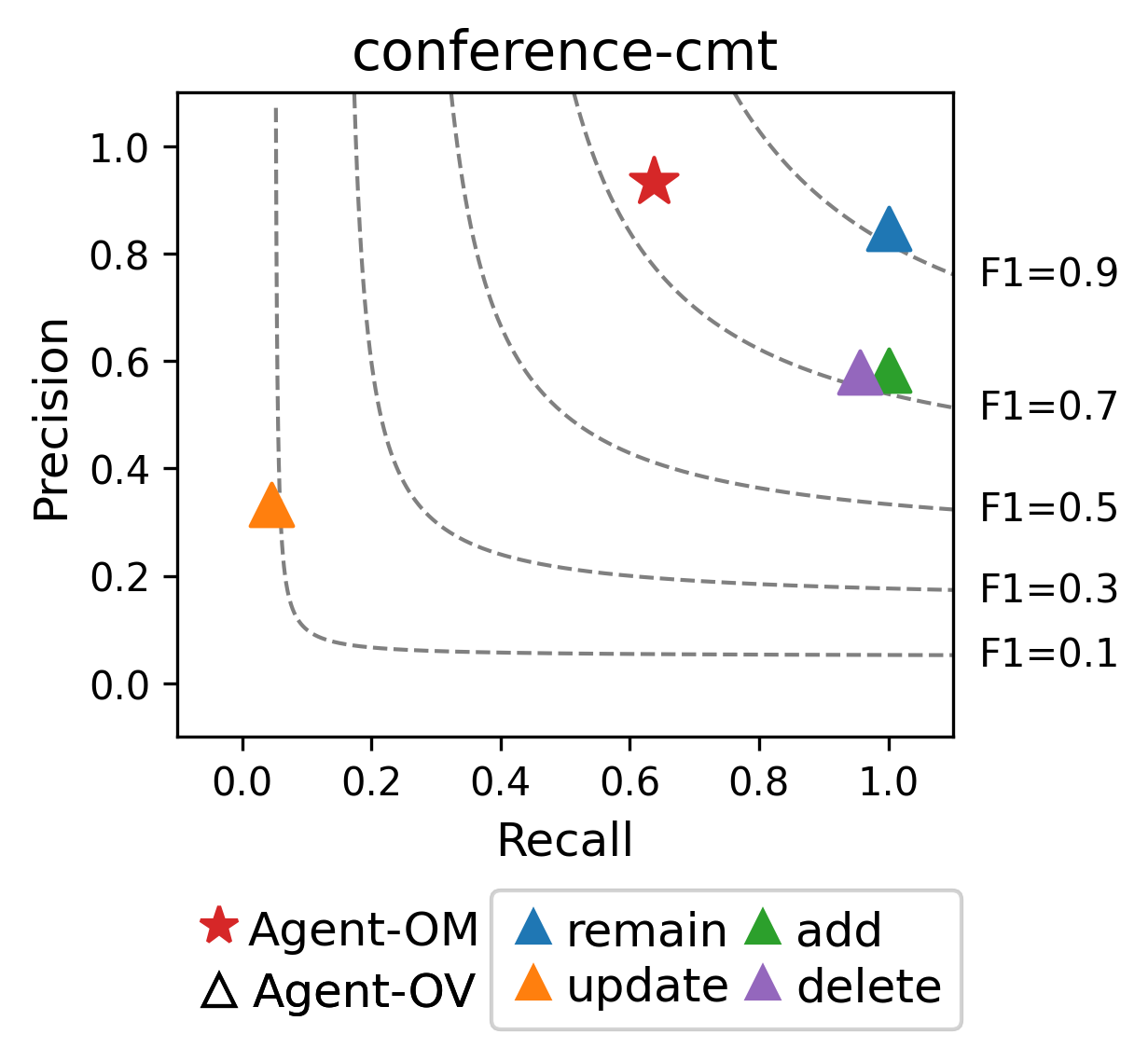}
\includegraphics[width=0.325\textwidth]{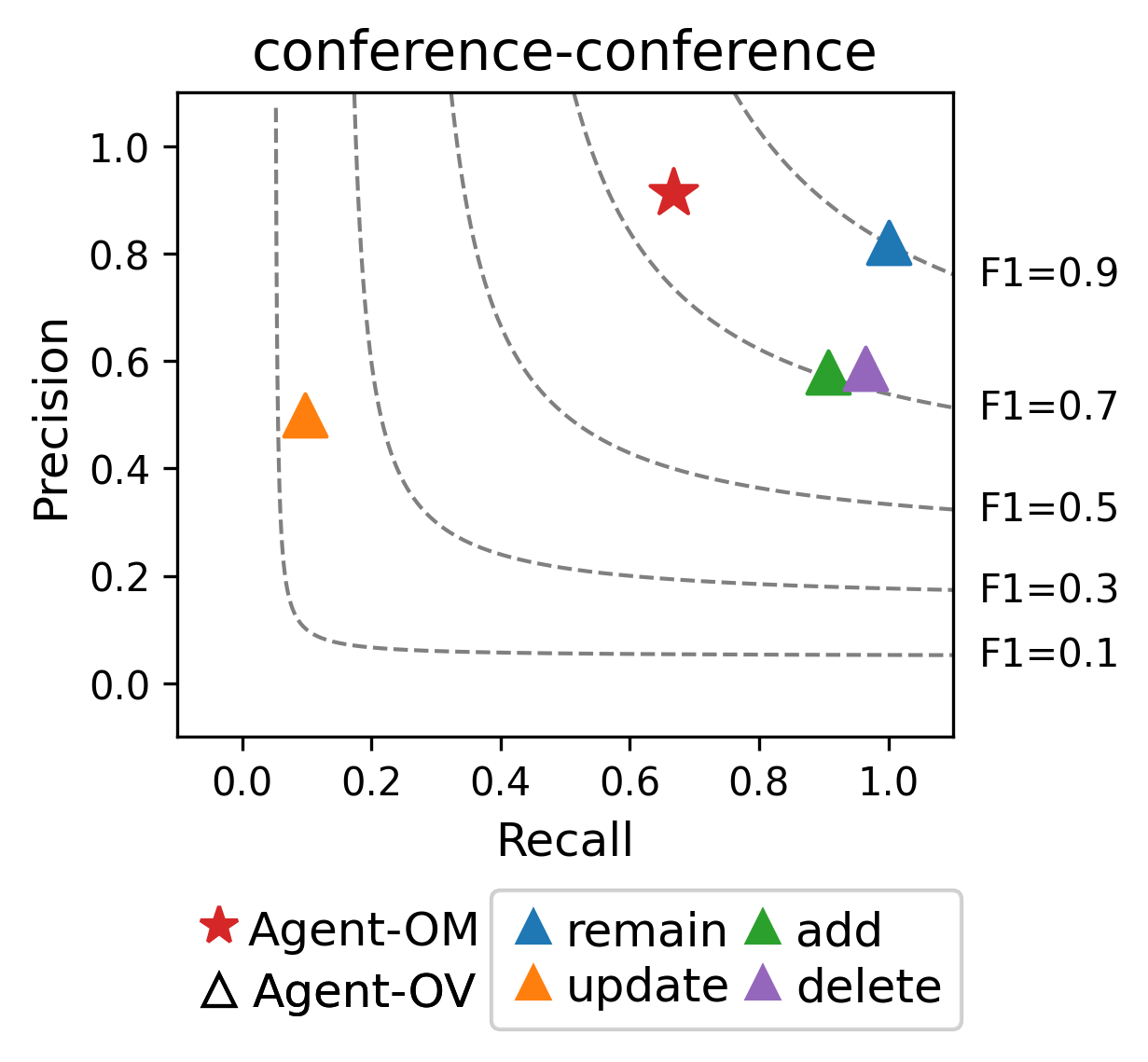}
\includegraphics[width=0.325\textwidth]{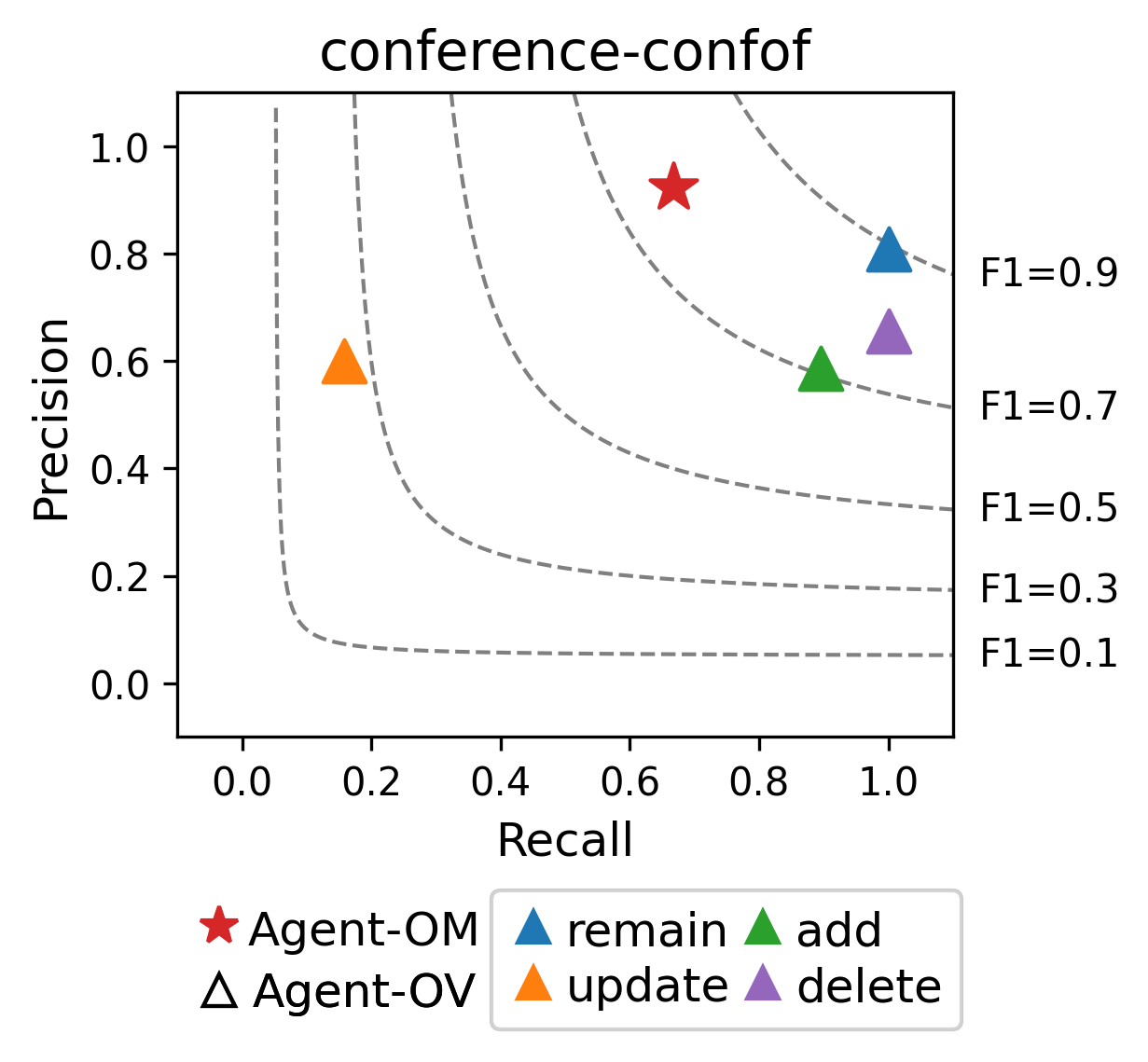}        \\
\includegraphics[width=0.325\textwidth]{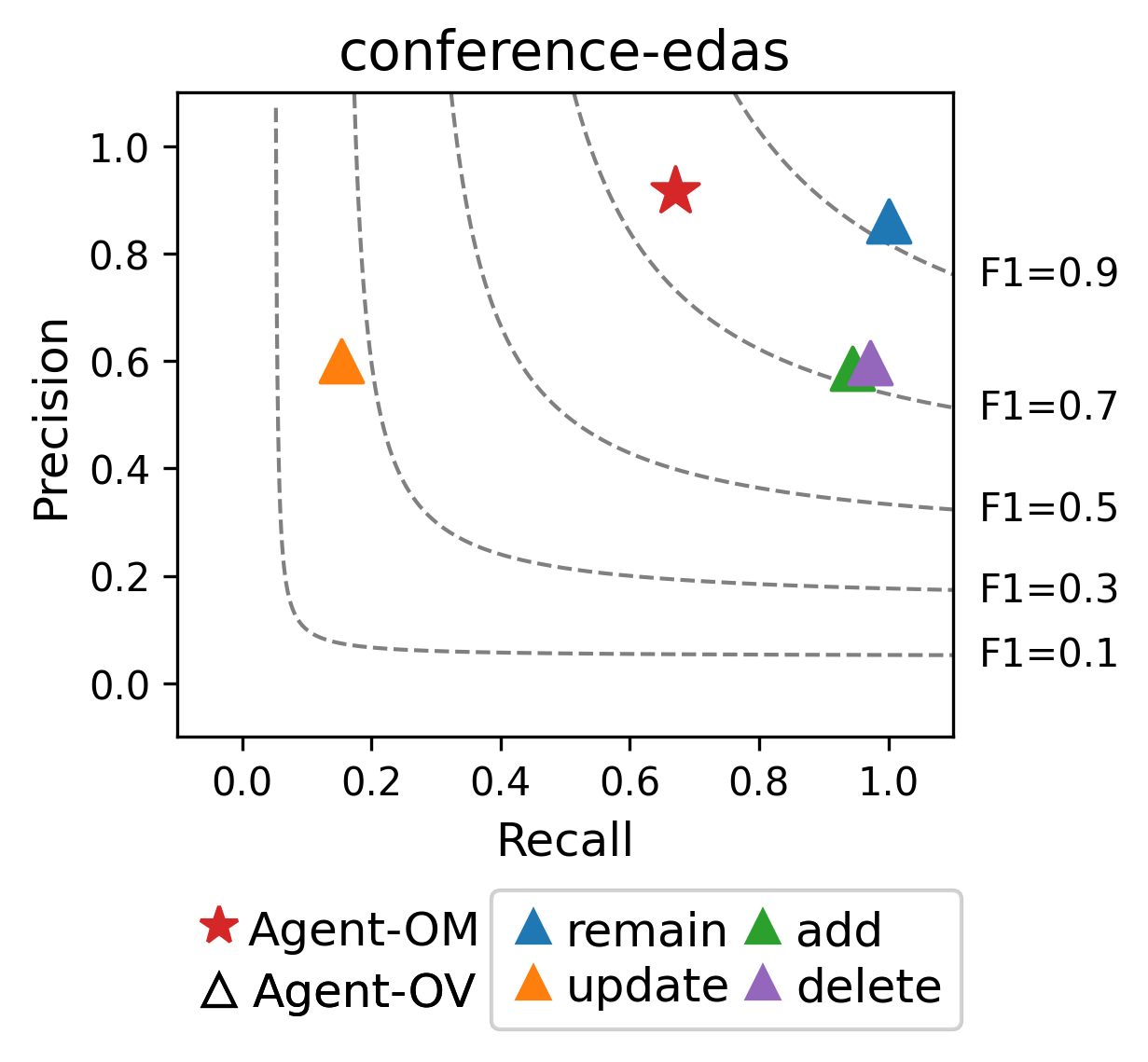}
\includegraphics[width=0.325\textwidth]{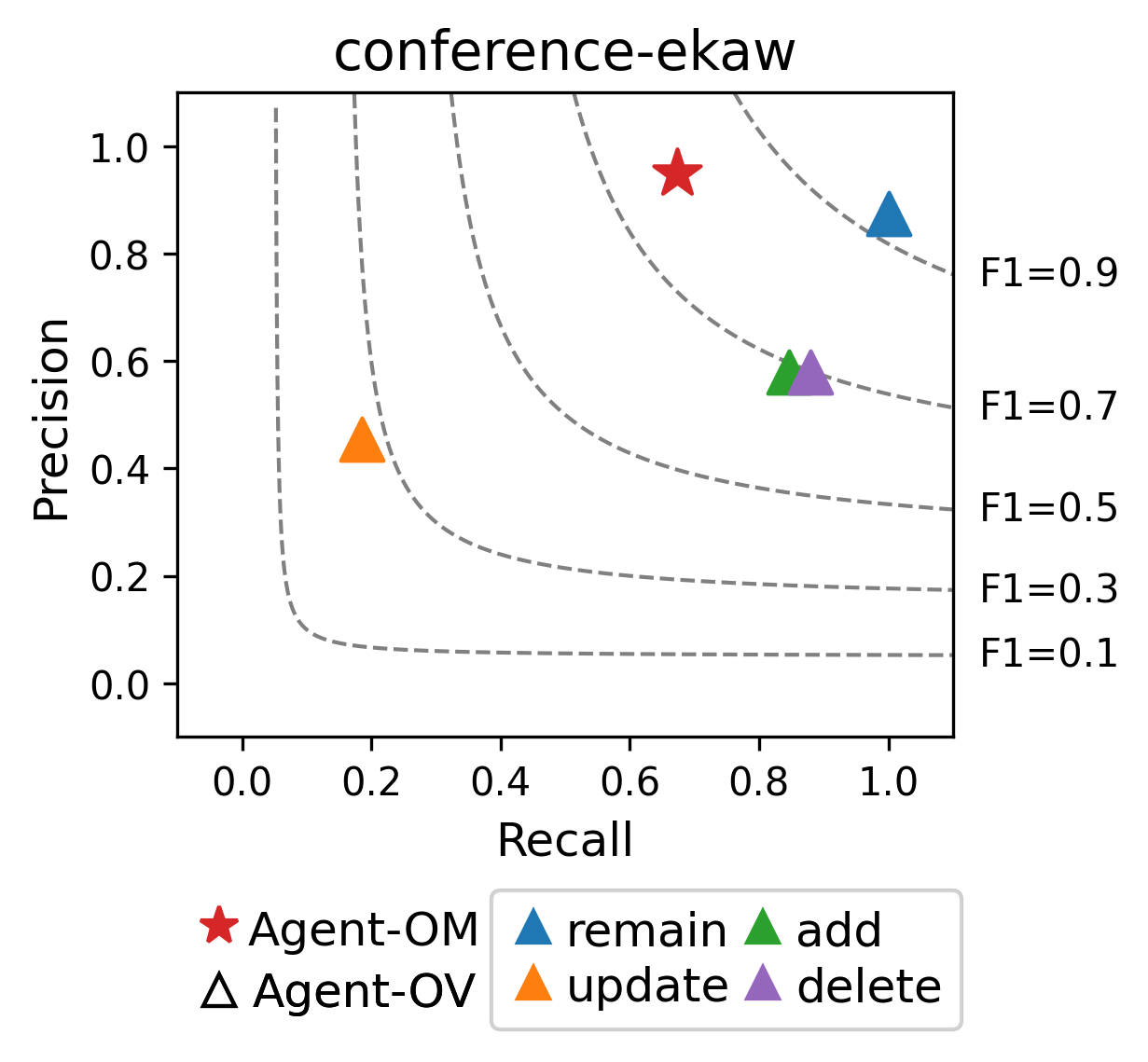}          \\
\includegraphics[width=0.325\textwidth]{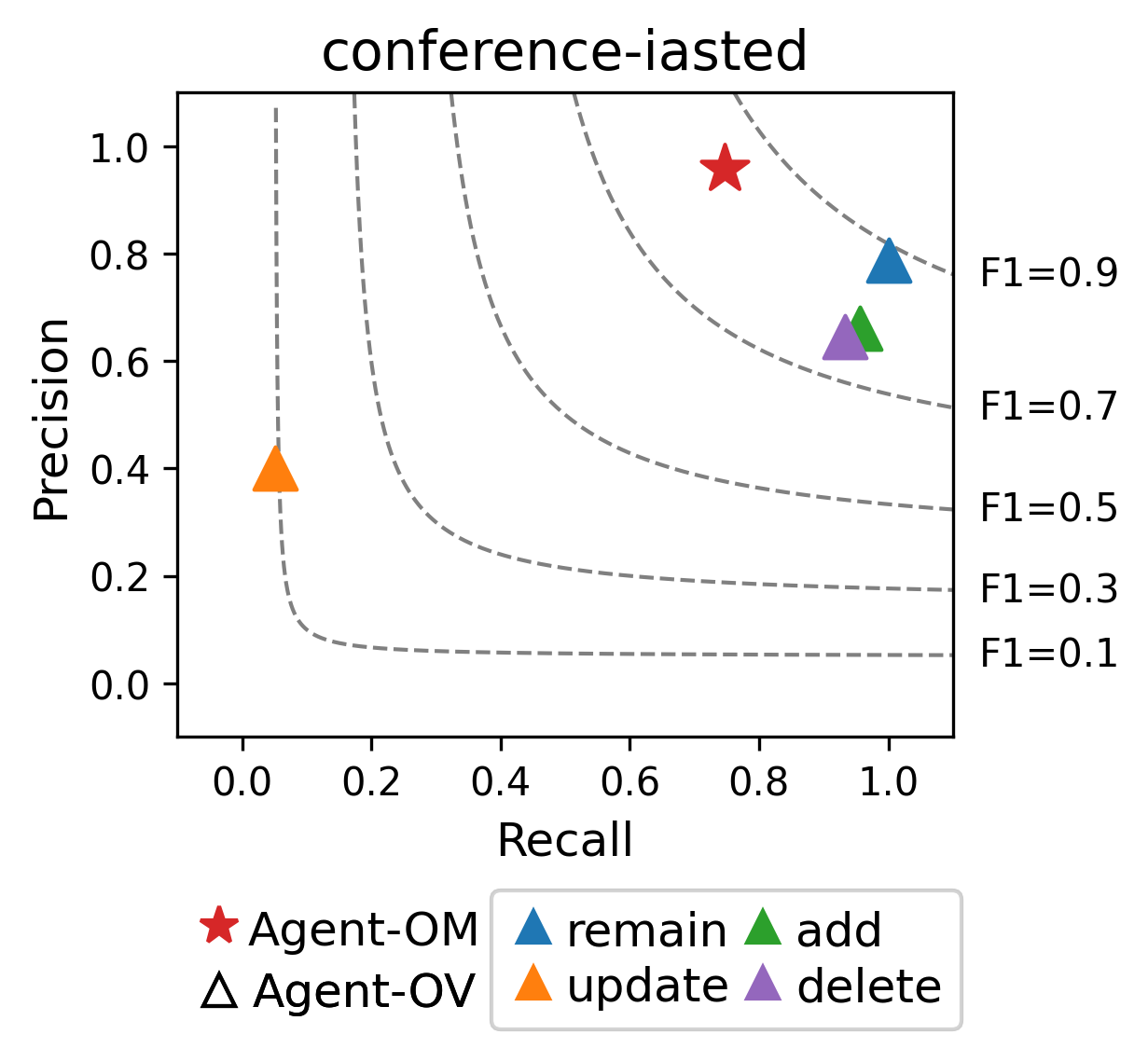}
\includegraphics[width=0.325\textwidth]{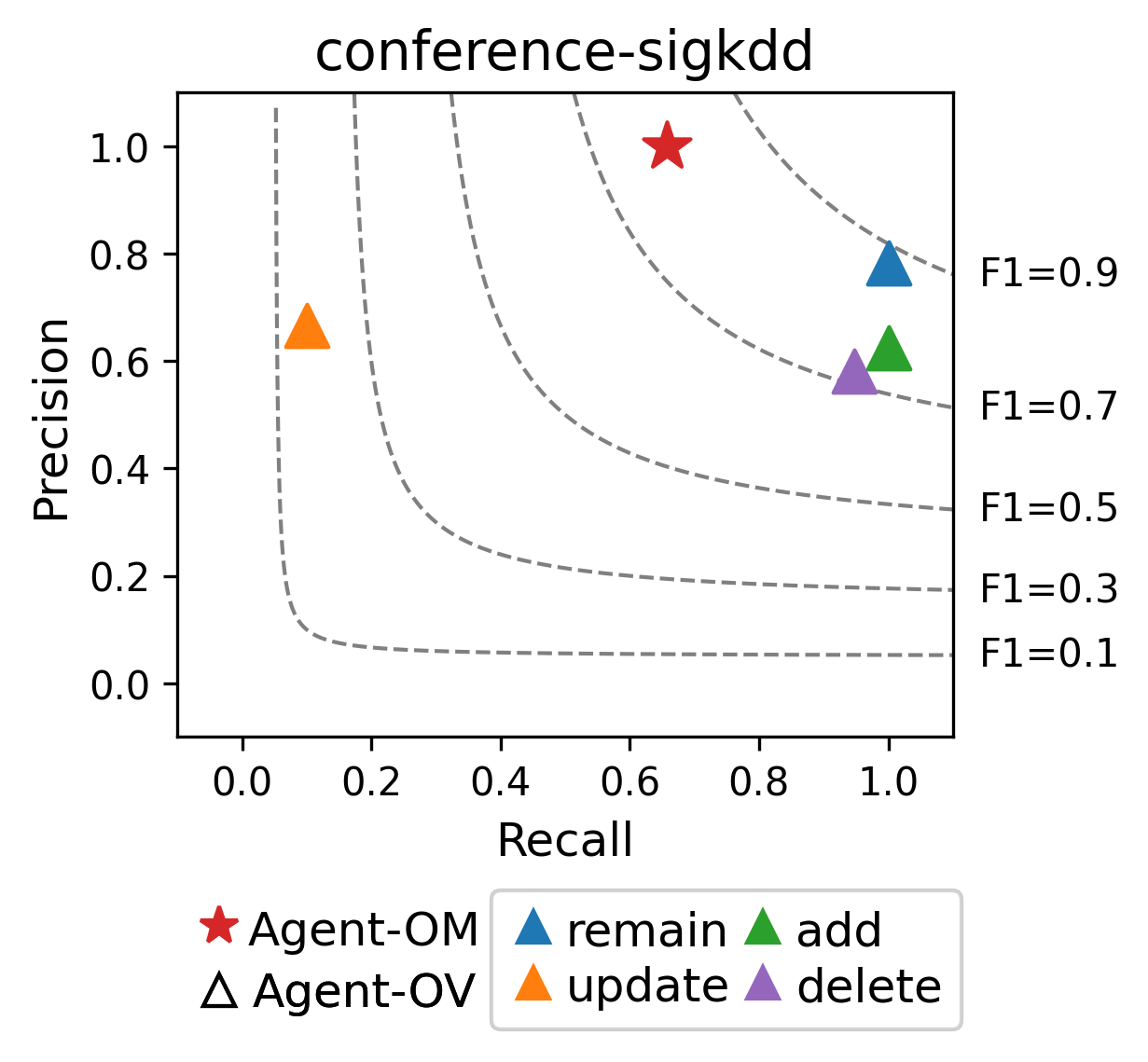}        \\
\caption{Evaluation of OAEI Conference Track on the OV testbed. Agent-OM, shown as a red star, determines only the single category of entity matches. The extended Agent-OV, shown as triangles, determines four different categories of entity matches (\textit{remain} or \textit{update}) and non-matches (\textit{add} or \textit{delete}).}
\label{fig: evaluation-conference}
\end{figure}

Figure~\ref{fig: evaluation-conference} shows the performance of Agent-OM and Agent-OV on the OAEI Conference Track. The results indicate that it is possible to unify the OM and OV tasks using the OM4OV framework. Both Agent-OM and Agent-OV can produce alignments with precision, recall, and F1 scores. However, we can see that Agent-OM consistently performs surprisingly well across all alignments. This is because the traditional OM system captures only matched entities between two ontologies, which in the OV task correspond to the \textit{remain} and \textit{update} entities. Because \textit{remain} entities typically comprise the major proportion of entities over two different versions, the correctness of \textit{remain} entities dominates the overall evaluation and other categories are de-emphasised. We believe this is a common situation that leads to misleading performance measurement when using an OM system in OV tasks.

Agent-OV overcomes the skewed measure in Agent-OM. By decomposing the measure into four sub-measures, we can observe more informative matching performance across different OV tasks. In general, we observe that the matching performance is highest in \textit{remain}, followed by \textit{add} and \textit{delete}, and relatively low in \textit{update}. This trend is consistent across different tracks and ontologies. The measures for \textit{remain} are typically very close to 100\%. The measures for \textit{add} and \textit{delete} are generally good. Our system uses LLMs as the backend, and LLMs generally have strong background knowledge to detect non-matched entities. The measures for \textit{update} show scope for improvement. This is because finding non-trivial alignments and appropriate similarity thresholds is not easy. We examine the false mappings produced by Agent-OV. Unlike OM tasks, we find that some of these identified false mappings are not actually false. Several examples are listed below with explanations.

\begin{bracketenumerate}
\item False mappings in OV can result from flawed ontology design choices, as shown here.
\begin{lstlisting}
(edas:ConferenceEvent, edas:Event, c)
(edas:ConferenceEvent, edas:AcademicEvent, c)
\end{lstlisting}
In the example, the reference shows \texttt{edas:ConferenceEvent} is updated to \texttt{edas:Event}, but the OV system predicts \texttt{edas:ConferenceEvent} is updated to \texttt{edas:AcademicEvent}. An event can be a non-academic event (e.g. a social event) or an academic event (e.g. a seminar). In the context of a research conference, a conference event is more likely to mean the latter. However, mapping to event still makes sense. Within one ontology, the meaning of two entities is too close to be distinguished and therefore leads to them being synonyms for each other. Ideally, we should avoid this type of ontology design. This example also demonstrates a unique benefit of using OM for OV, which could potentially assist in ontology design.
\item False mappings in OV can create ambiguous ``equivalent'' relationships, as shown here.
\begin{lstlisting}
(confof:dealsWith, confof:hasSubjectArea, c)
(confof:dealsWith, confof:reviewes, c)
\end{lstlisting}
In the example, the reference shows \texttt{\small confof:dealsWith} is updated to \texttt{\small confof:hasSubjectArea}, but the OV system predicts \texttt{confof:dealsWith} is updated to \texttt{confof:reviewes}. This is not necessarily wrong, but it interprets ``deal with'' differently. It is vital to notice that the term ``equivalent'' in OV is weaker than that in OM. OV allows for roughly ``equivalent'' mappings. The entities mapped in OV can slightly alter their meanings in response to real-world changes in the domain or evolution of natural language.
\item False mappings in OV can arise from different similarity thresholds, as shown here.
\begin{lstlisting}
(sigkdd:Start_of_conference, None, 0.90)
(sigkdd:Start_of_conference, sigkdd:startDate, 0.80)
\end{lstlisting}
In the example, if $similarity\_threshold = 0.90$, \texttt{sigkdd:Start\_of\_conference} in $O$ will be assigned to \textit{delete} entities as it does not have matching entities; if $similarity\_threshold = 0.80$, \texttt{sigkdd:Start\_of\_conference} in $O$ will be assigned to \textit{update} entities as it has a matching entity \texttt{sigkdd:startDate} in $O'$. Both results are valid because defining the boundary between matching and non-matching is context- and application-dependent. For example, the similarity threshold could be relatively higher in the biomedical domain to ensure that each term is unique, whereas the similarity threshold in the conference domain could be relatively lower to improve the interoperability of terminologies used in research conferences.
\end{bracketenumerate}

\begin{figure}[htbp]
\centering
\includegraphics[width=0.325\textwidth]{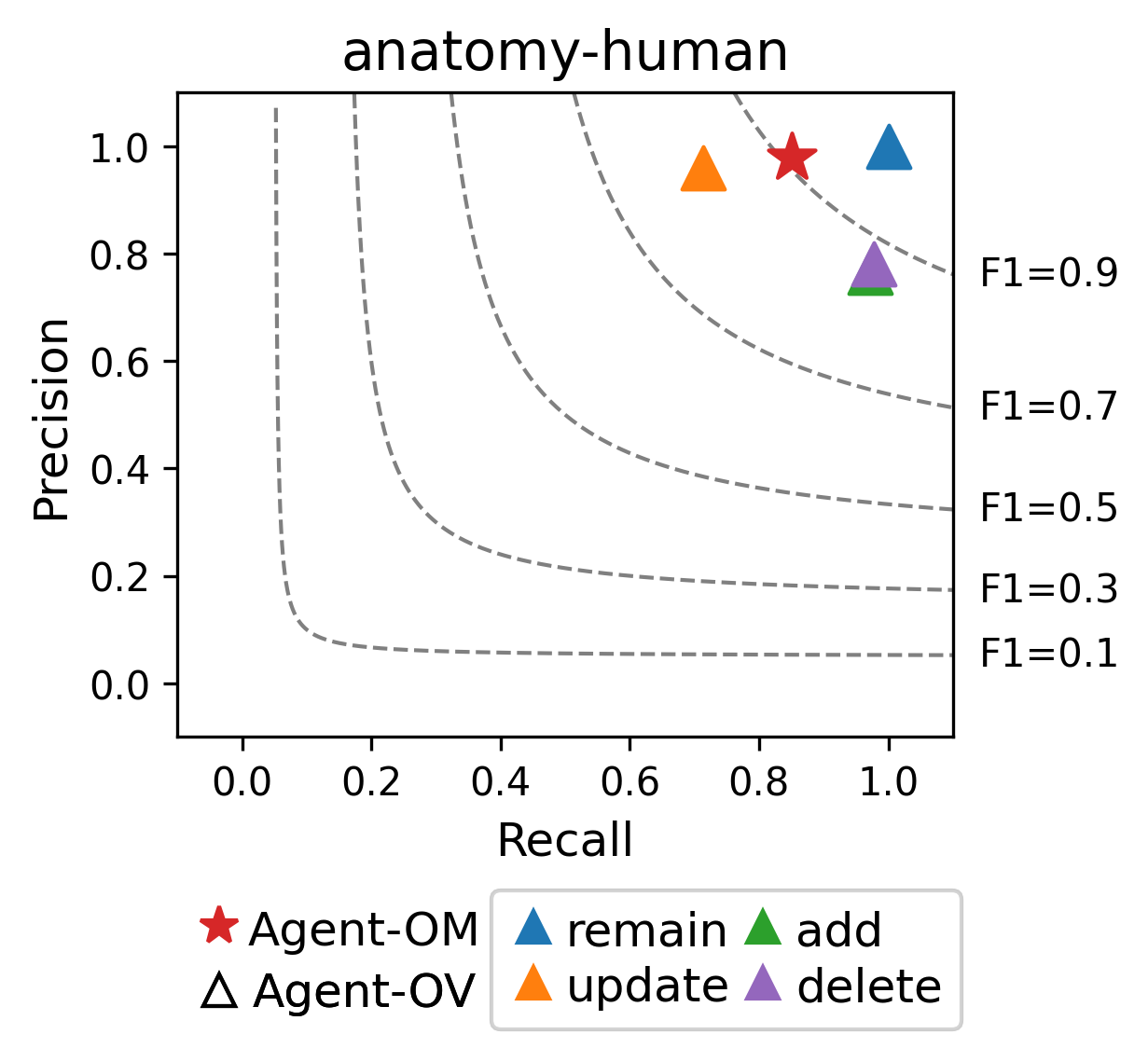}
\includegraphics[width=0.325\textwidth]{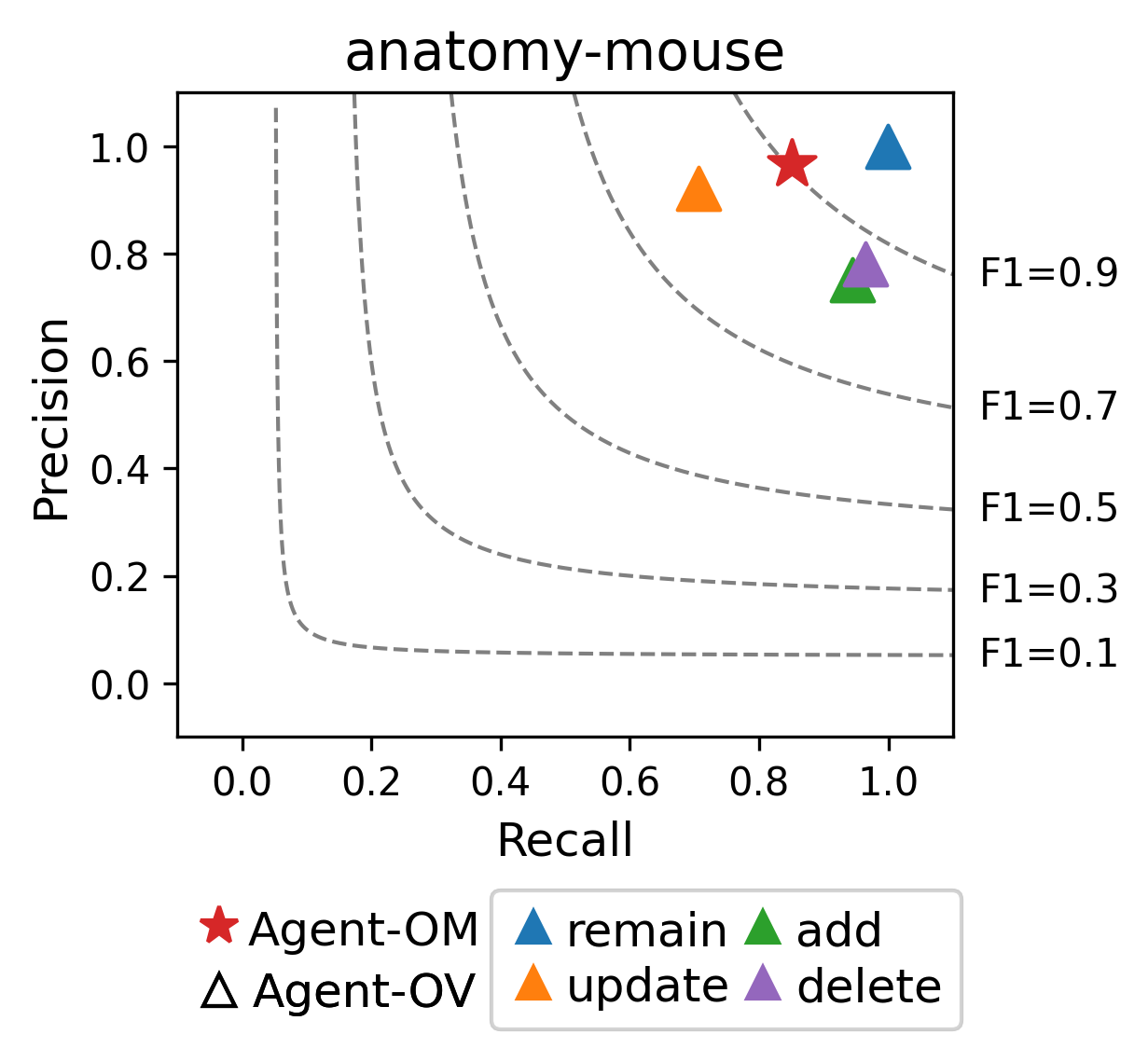}    \\
\includegraphics[width=0.325\textwidth]{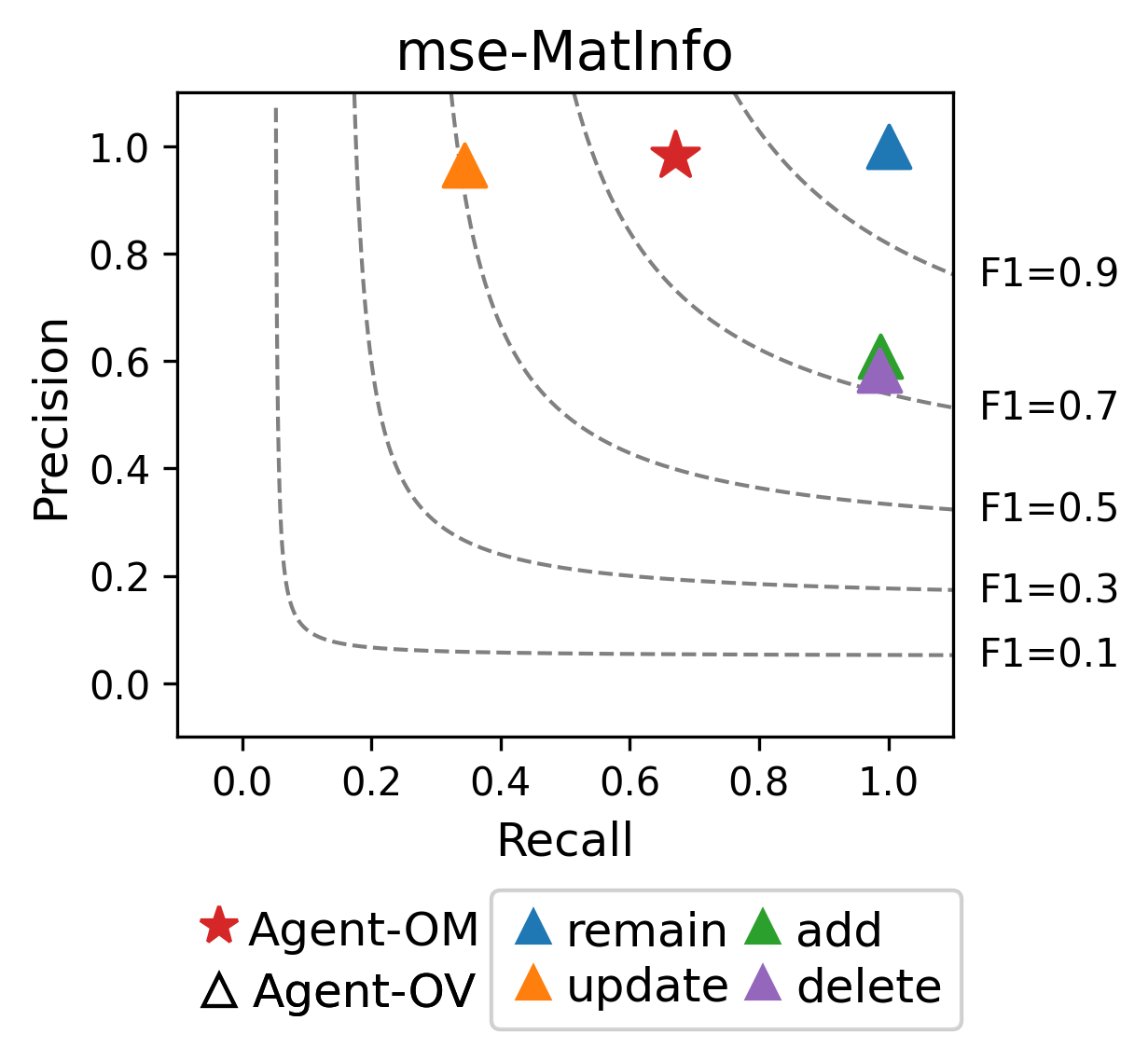}
\includegraphics[width=0.325\textwidth]{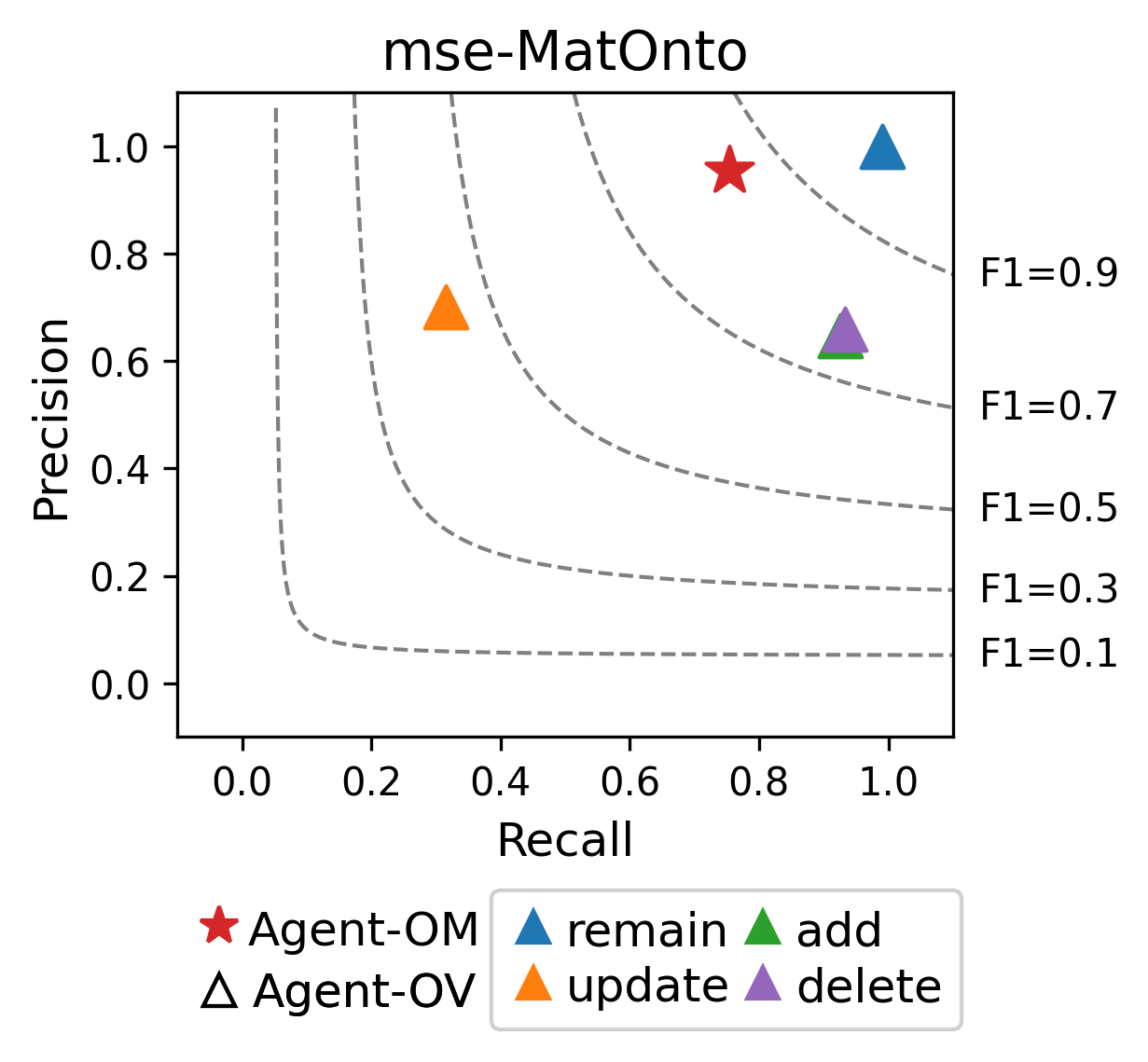}      \\
\caption{Evaluation of OAEI Anatomy and MSE Tracks on the OV testbed has shown similar trends.}
\label{fig: evaluation-other}
\end{figure}

\begin{figure}[!t]
\includegraphics[width=1\linewidth]{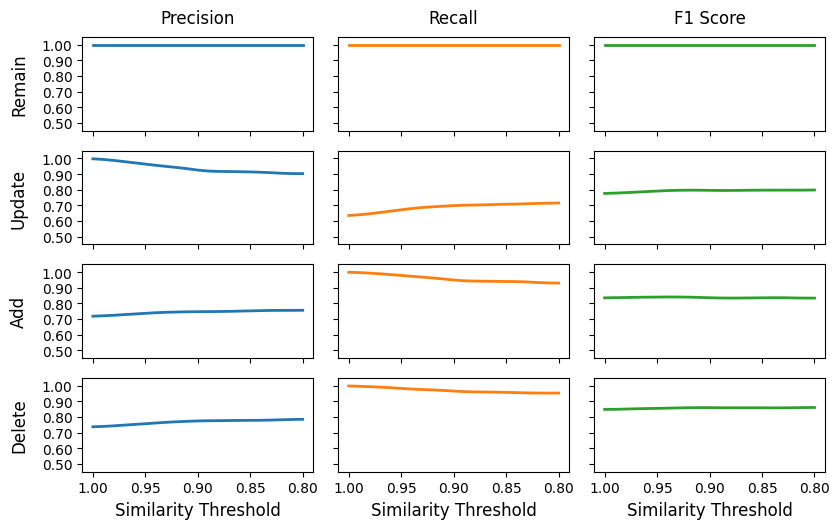}
\caption{Evaluation varying similarity thresholds in Agent-OV. We apply a 1-D Gaussian filter for smoothing. The standard deviation for a Gaussian kernel is 1 (i.e. $sigma=1$). We can see that the similarity threshold has no effect on detecting \textit{remain} entities in the top row, but affects the detection of \textit{update}, \textit{add}, and \textit{delete} entities.}
\label{fig: evaluation-similarity-without-cr}
\end{figure}

We extend our analysis to other OAEI tracks and varying similarity thresholds. Figure~\ref{fig: evaluation-other} shows that the trends are consistent with other medium and large OAEI tracks evaluated on the OV testbed. Additionally, we observe a longer computation time in the Anatomy Track. Although Agent-OV has an optimisation module for the matching candidate selection process (inherited from Agent-OM), it is still insufficient for some OV tasks. There is a need to optimise the framework for OV in large-scale ontologies. Figure~\ref{fig: evaluation-similarity-without-cr} studies the effect of varying similarity thresholds for the Mouse ontology in Agent-OV. We change the similarity threshold from 1.00 to 0.80 to evaluate its effect on matching performance. While OV entities are classified into four different categories, each category requires a corresponding sub-measurement. Similar to OM tasks, changes in hyperparameter settings in OV tasks lead to a trade-off between precision and recall. Moreover, the hyperparameter settings also influence the sub-measures. We find that the similarity threshold has no effect on detecting \textit{remain} entities, but significantly affects the detection of \textit{update} entities and slightly affects the detection of \textit{add} and \textit{delete} entities. Detecting \textit{update} entities and detecting \textit{add} and \textit{delete} entities are negatively correlated, as is expected because a near-miss \textit{update} alignment will fall back to a pair of \textit{non-match} entities, one from each ontology. Therefore, lower similarity thresholds can result in more \textit{update} entities being detected, while higher similarity thresholds may find more \textit{add} and \textit{delete} entities.

\subsection{Discussion}
\label{sec: discussion}

The experimental results have shown that OM systems can be reused for OV tasks, but they have limitations and require necessary extensions. The following key points need to be taken into account when reusing OM systems for OV tasks.
\begin{bracketenumerate}
\item \textbf{Skewed Measurement:} OM systems exhibit skewed measurement of OV tasks because unchanged \textit{remain} entities dominate in practice. OM measures are reusable, but these measures need to be extended into four sub-measures for \textit{add}, \textit{delete}, \textit{remain}, and \textit{update} performance. Within each sub-measure, the inherent precision-recall trade-off still holds. Across different sub-measures, they are not independent. This is because each change in a part of an alignment in one category will cause other categories to change accordingly. For example, if a new mapping is found in $A\odot$ or $A\otimes$, then the number of mappings in $A\oplus$ and $A\ominus$ will be reduced by one each, and vice versa.
\item \textbf{Update Pitfalls:} OM systems are sensitive to changes in similarity thresholds and backend LLMs. For OV tasks, only \textit{update} entities are sensitive to these changes, while \textit{remain}, \textit{add}, and \textit{delete} are relatively stable across these hyperparameter settings. Poor \textit{update} performance is common, while tuning the similarity threshold trades precision for recall without resolving the underlying difficulty.
\item \textbf{Disputed False Mappings:} OM systems used for OV tasks can produce mappings that are false with respect to the ground truth reference, but that are true with respect to expert review. This can be caused by flawed ontology design choices, ambiguous ``equivalent'' relationships, or inappropriate settings of the similarity threshold. This is difficult for OM systems to recognise and generally requires human effort.
\end{bracketenumerate}

\section{OM4OV Optimisation}
\label{sec: optimisation}

Having established the current pitfalls of OM4OV, we explore optimisations for the framework. Often, ontology creators provide cross-references to alternative ontologies to enhance interoperability and facilitate integration. For example, a cross-reference between the Mouse ontology and the Human ontology is provided in the OAEI Anatomy Track. Reusing the cross-references developed for OM tasks, we propose a novel mechanism to reduce the number of matching candidates and also to improve overall OV performance.

\begin{figure}[htbp]
\centering
\includegraphics[width=0.6\linewidth]{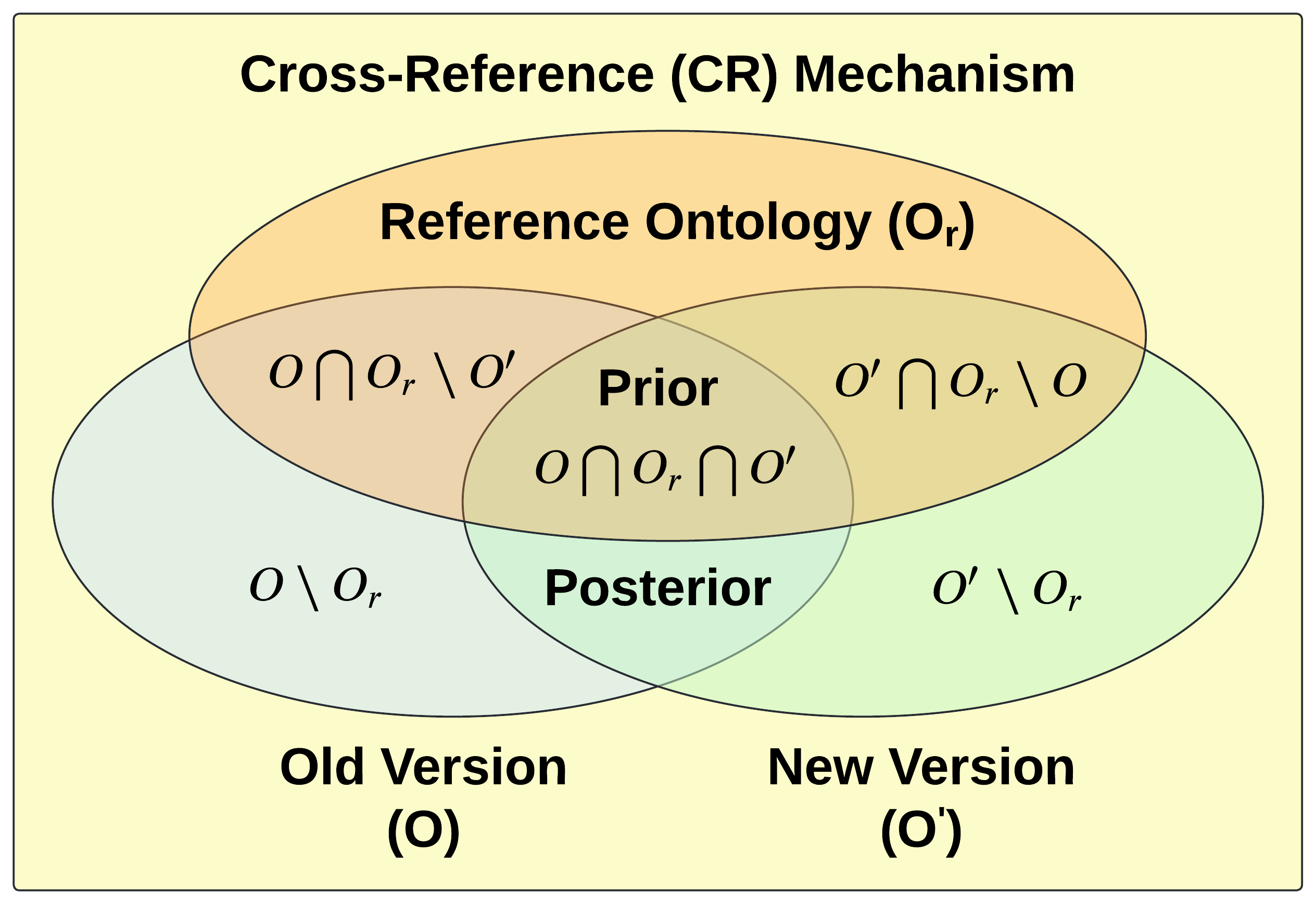}
\caption{Illustration of the cross-reference (CR) mechanism.}
\label{fig: om4ov-extend}
\end{figure}

Figure~\ref{fig: om4ov-extend} illustrates the cross-reference (CR) mechanism used in the OM4OV framework. We can see that, without using the cross-reference $O_r$, the matching candidates cover the range of $O\cup O'$. This number can be significantly reduced by removing prior matches (i.e. $O \cap O_r \cap O'$) and non-matches (i.e. $ O\cap O_r \setminus O'$ and $O' \cap O_r \setminus O$). The prior matching will be incorporated into the final alignment, while the known non-matching regions will be removed entirely in the subsequent OV process. The OV process then only determines the posterior alignment.

In practice, the prior alignment usually contains a large number of \textit{remain} entities and a small number of \textit{update} entities. Matching performance is also enhanced by utilising these established mappings. Since prior matches are inferred from OM references validated by domain experts, they provide a solid ground truth for alignment within a specific domain. Removing known non-matches from the OV process also simplifies posterior alignment detection. The optimised OM4OV framework using the CR mechanism is defined as follows.
\begin{definition}
Given a reference ontology ($O_r$), an old version of an ontology ($O$) and a new version of the same ontology ($O'$), two cross-references between $O$ and $O_r$ ($R_{or}$) and between $O'$ and $O_r$ ($R_{o'r}$) are defined respectively as:
\begin{equation}
\begin{aligned}
R_{or}  &= \{(e_1, e_3, c) \mid e_1 \in O, e_3 \in O_r, s \leq c \leq 1 \mbox{ and there is no } (e_1,e_3,c')\mbox{ with } c'> c\}    \\
R_{o'r} &= \{(e_2, e_3, c) \mid e_2 \in O', e_3 \in O_r, s \leq c \leq 1 \mbox{ and there is no } (e_2,e_3,c')\mbox{ with } c'> c\}   \\
\end{aligned}
\end{equation}
\end{definition}

We can use $R_{or}$ and $R_{o'r}$ to infer some known mappings between $O$ and $O'$ before performing OV. We call these mappings a \textit{prior} alignment ($A_{\pi}$). After subsequently performing OV, we have our \textit{posterior} alignment ($A_{\pi^*}$). In this setting, $A_{match}$ in OV can be decomposed into two parts: $A_{match} = A_{\pi} \cup A_{\pi^*}$. $A_{\pi}$ can be directly inferred from the two cross-references $R_{or}$ and $R_{o'r}$. By assuming the equivalence relation as extracted from the two cross-references is transitive, we have that $e_1 \in O$, $e_2 \in O'$, and $e_3 \in O_r$, if $e_1 \equiv e_3$ and $e_2 \equiv e_3$ then $e_1 \equiv e_2$. $A_{\pi^*}$ aims to detect missing mappings from the cross-reference. None of these mappings would come from any pairwise intersection of $O$, $O_r$, and $O'$ because $O \cap O_r \setminus O'$ and $O' \cap O_r \setminus O$ are pre-defined as non-matched entities, and the matched entities in $O\cap O_r \cap O'$ have already been captured in the $A(\pi)$. As a result, $A_{\pi^*}$ can be defined within a smaller scope.
\begin{definition}
\textit{Prior} alignment ($A_{\pi}$) and \textit{posterior} alignment ($A_{\pi^*}$) are defined as:
\begin{equation}
\begin{aligned}
A_{\pi} &= R_{or} \cap R_{o'r} = \{(e_1, e_2, c) \mid e_1, e_2 \in O \cap O_r \cap O', s \leq c \leq 1\}    \\
A_{\pi^*} &= \{(e_1, e_2, c) \mid e_1 \in O\setminus O_r, e_2 \in O' \setminus O_r, s \leq c \leq 1\}       \\
\end{aligned}
\end{equation}
\end{definition}

An ontology can have multiple cross-references available. In such cases, the \textit{prior} reference becomes the union of all known cross-references ($R_{or1}$ ... $R_{orn}$), and the ontology used in the \textit{posterior} alignment ($O_{ra}$) becomes the union of all reference ontologies ($O_{r1}$ ... $O_{rn}$).
\begin{definition}
$A_{\pi}$ and $A_{\pi^*}$ in multiple cross-references can be formulated as:
\begin{equation}
\begin{aligned}
A_{\pi} &= (R_{or1} \cap R_{o'r1}) \cup (R_{or2} \cap R_{o'r2}) \cup ...\cup (R_{orn} \cap R_{o'rn}) \\
A_{\pi^*} &= \{(e_1, e_2, c) \mid e_1 \in O\setminus O_{ra}, e_2 \in O' \setminus O_{ra}, c \geq s\} \ \text{where} \ O_{ra} = O_{r1} \cup O_{r2} \cup... O_{rn} \\
\end{aligned}
\end{equation}
\end{definition}

\begin{figure}[!t]
\includegraphics[width=1\linewidth]{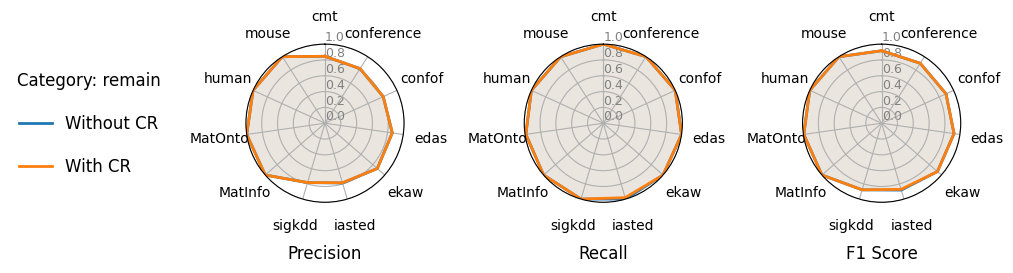}
\includegraphics[width=1\linewidth]{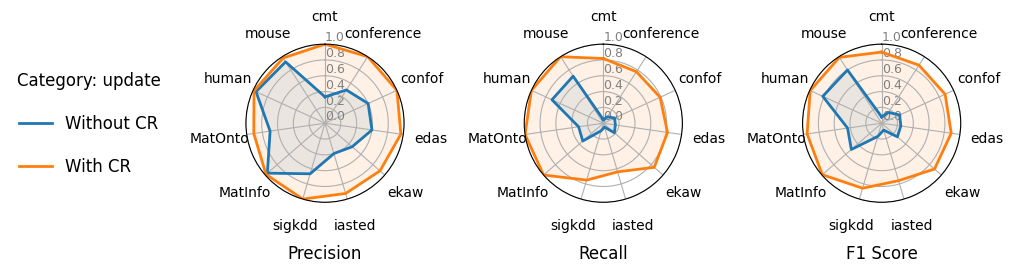}
\includegraphics[width=1\linewidth]{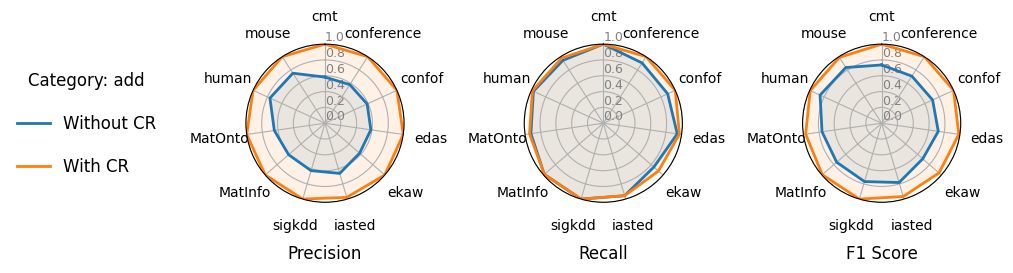}
\includegraphics[width=1\linewidth]{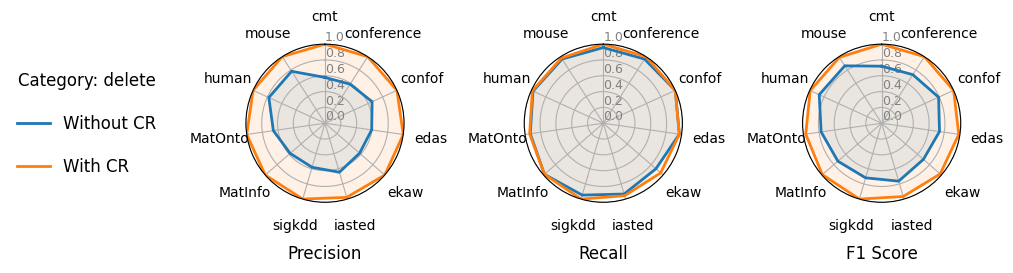}
\caption{Evaluation of the cross-reference (CR) mechanism on the OV testbed. Each radar chart shows measurements (precision, recall, and F1 score) for a change category over 11 datasets analysed around the perimeter, both with (\textit{orange}) and without (\textit{blue}) the CR mechanism applied.}
\label{fig: evaluation-cr}
\end{figure}

\begin{figure}[!t]
\includegraphics[width=1\linewidth]{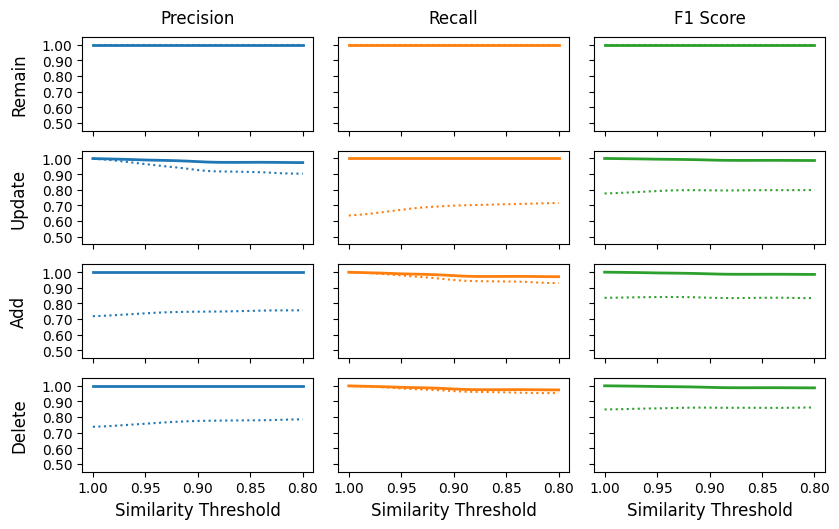}
\caption{Evaluation of the cross-reference (CR) mechanism on different similarity thresholds. The solid lines contribute to OV performance with the CR mechanism, while the dotted lines contribute to OV performance without the CR mechanism. We apply a 1-D Gaussian filter for smoothing. The standard deviation for a Gaussian kernel is 1 (i.e. $sigma=1$).}
\label{fig: evaluation-similarity-with-cr}
\end{figure}

Figure~\ref{fig: evaluation-cr} compares the OV performance with and without the CR mechanism on the same OV testbed that we analysed in Section~\ref{sec: evaluation}. We can see that the CR mechanism significantly enhances Agent-OV's ability to detect \textit{update} entities and slightly improves its performance in detecting \textit{add} and \textit{delete} entities, with no effect on \textit{remain} entities. More specifically, performance improvement is mainly attributed to recall for \textit{update} entities, whereas it is more evident in precision for \textit{add} and \textit{delete} entities. Figure~\ref{fig: evaluation-similarity-with-cr} compares the effects of different similarity thresholds for the Mouse ontology versioning. The results of the experiment show that the CR mechanism is less sensitive to the similarity threshold than the original OM4OV framework in Figure~\ref{fig: evaluation-similarity-without-cr}, suggesting that it can be helpful in scenarios where the optimal similarity threshold is unclear or difficult to determine.

\section{OM4OV in Real-World Applications}
\label{sec: application}

Brick Schema~\cite{balaji2016brick} is a popular ontology used in the building industry. Introduced in 2016, Brick Schema has undergone several major version changes, including URI changes, schema language changes, and ontology harmonisation (combining Brick Schema with RealEstateCore~\cite{hammar2019realestatecore} and Project Haystack~\cite{john2020project}). We use the change logs from the official GitHub repository to provide near-ground-truth alignments for evaluation. We are not confident about the reliability of the manually-generated logs as we observe instances where the log entry for a change conflicts with the difference observed in the ontology itself. We demonstrate versioning performance over three distinct major version updates: from v1.1.0 to v1.4.0, v1.2.0 to v1.4.0, and v1.3.0 to v1.4.0. We have collapsed multiple human-identified minor version changes into the corresponding major version changes for our experiments to consider more complex version changes than are evident in successive minor versions. We filter out classes and properties that are used in Brick Schema conventions but that user experts determined to be unimportant for version tracking and hence have not been captured in change logs. For example, ``Tag'', ``NodeShape'', and ``PropertyShape'' are excluded from the matching candidates.

\begin{figure}[!t]
\centering
\includegraphics[width=0.325\textwidth]{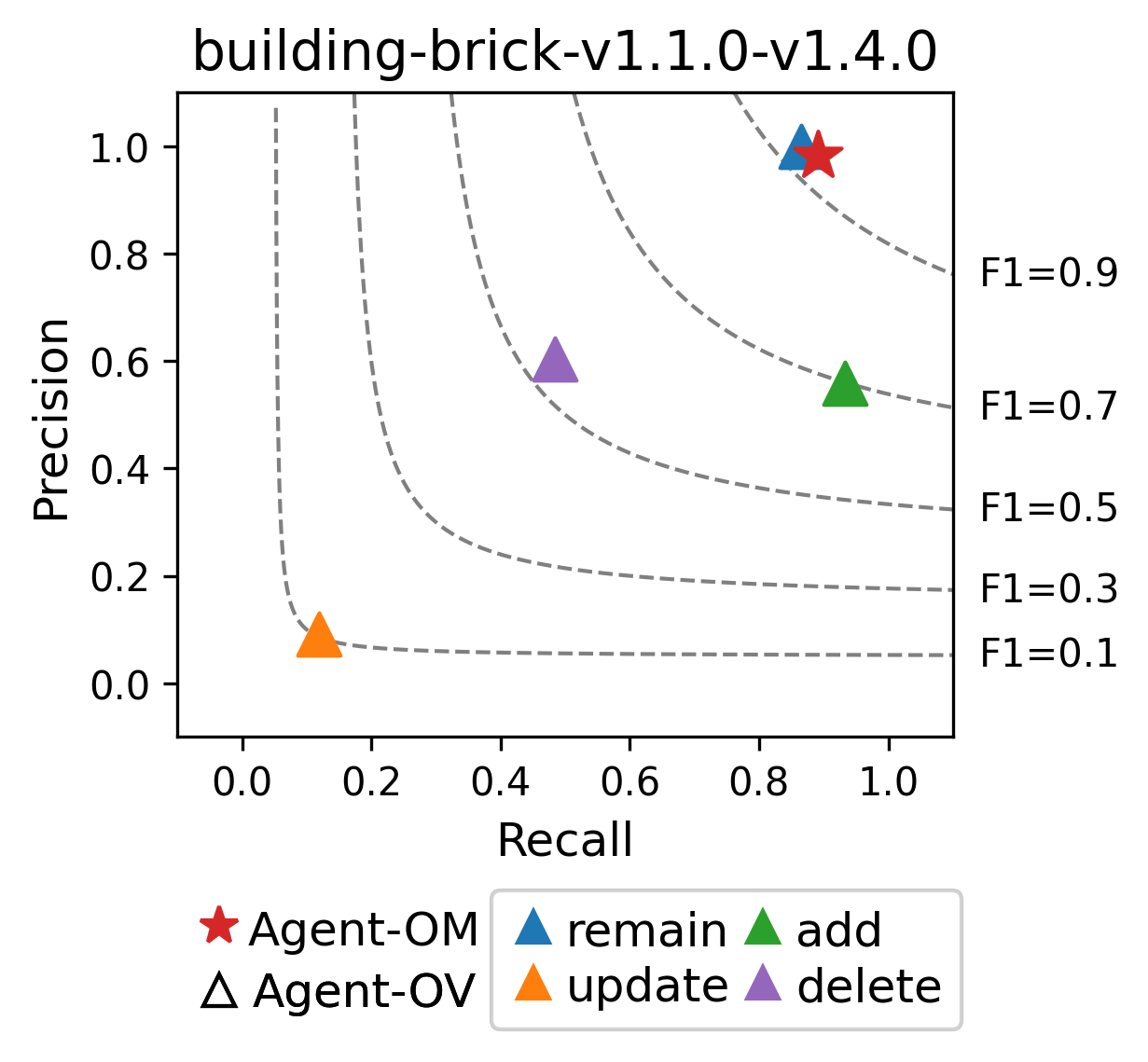}
\includegraphics[width=0.325\textwidth]{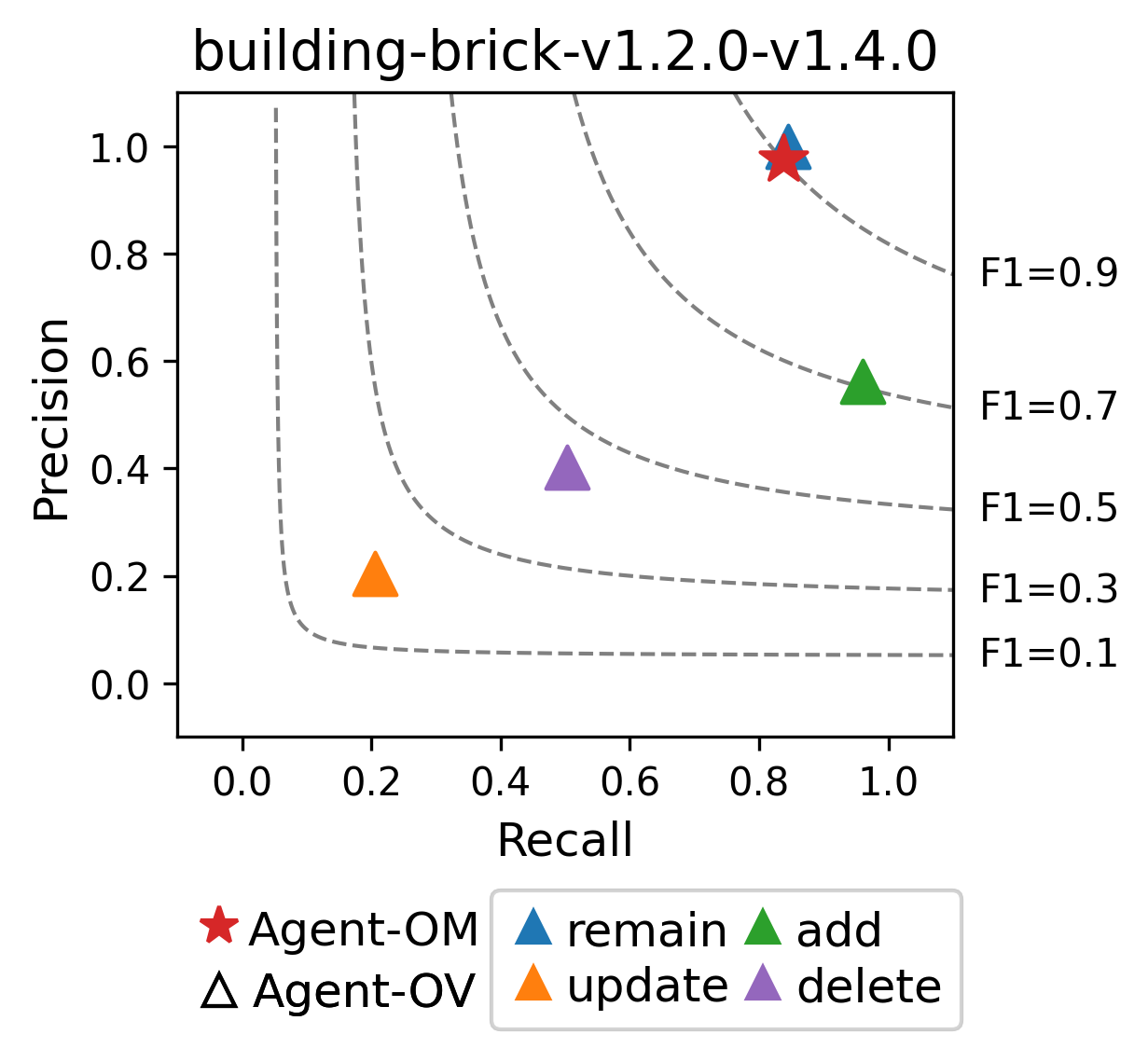}
\includegraphics[width=0.325\textwidth]{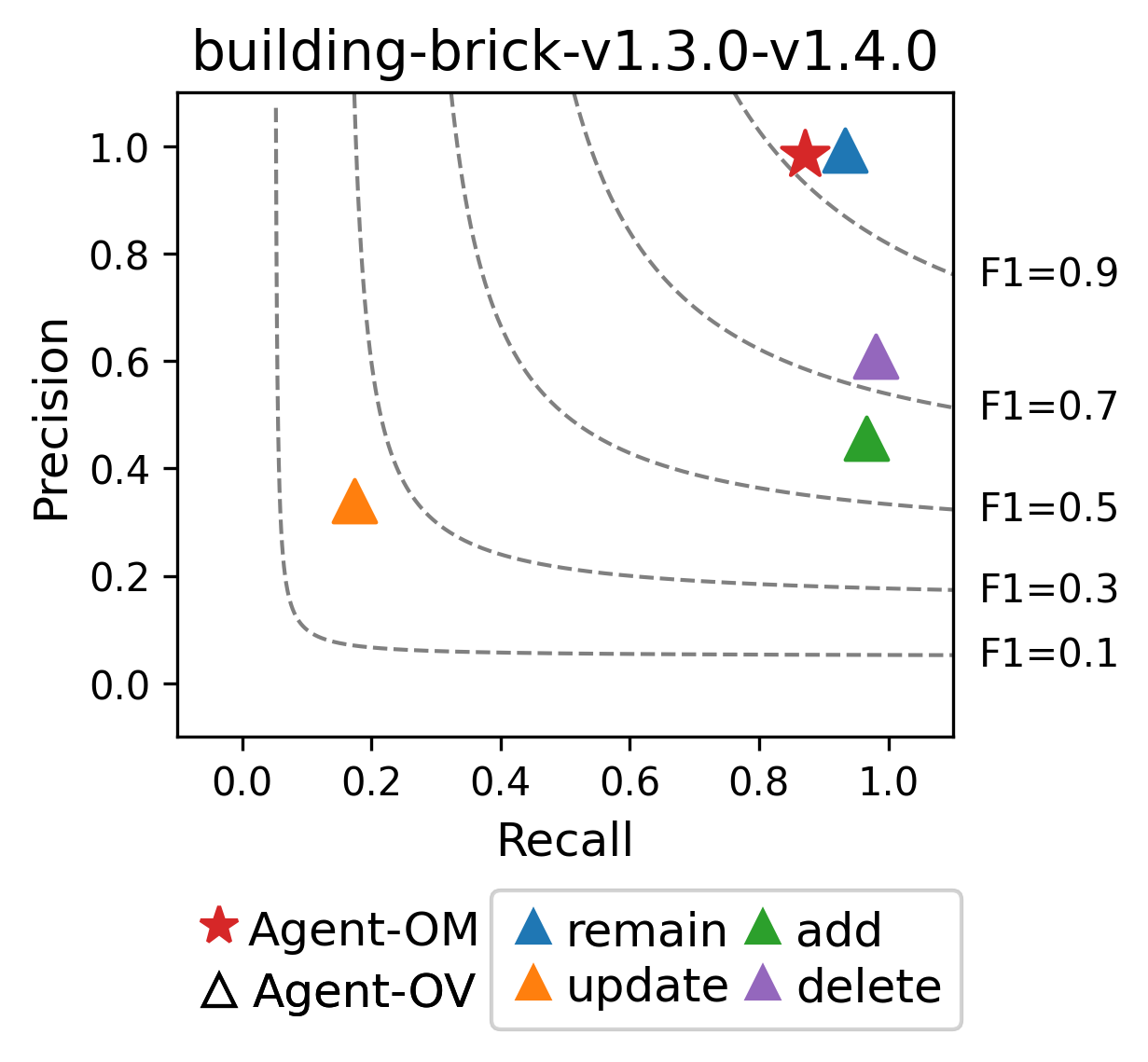}
\caption{Evaluation of the OM4OV framework tracking Brick Schema version changes.}
\label{fig: evaluation-brick}
\end{figure}

Figure~\ref{fig: evaluation-brick} shows that our OM4OV implemented in the OM system is more precise than our previous OM-only system in tracking changes across different versions of Brick Schema. Similar trends are observed in skewed measurements, update pitfalls, and disputed false mappings. We are unable to empirically evaluate our cross-reference mechanism with Brick Schema. Unfortunately, while Brick Schema offers high-level alignments with RealEstateCore and Project Haystack, these alignments are highly incomplete, addressing only few classes and properties. Further, they do not indicate versions of the ontologies to which they apply. We have been unable to locate an alternative ontology that offers the data needed for evaluating our cross-reference mechanism. We hope our work contributes to the quality improvement of OV so that building high-quality versioned reference alignments becomes standard industry practice. At that time, automated system updates with respect to ontology version changes would become feasible, and we could evaluate our cross-reference mechanism on real-world data.

\section{Conclusion}
\label{sec: conclusion}

In this paper, we systematically analyse similarities and differences between OM and OV tasks and provide formal definitions and task formulations for OM4OV. The experimental results from Agent-OM and Agent-OV demonstrate that the OM4OV framework is workable, though it presents several limitations. To tackle these issues, we propose a novel optimisation method that uses a cross-reference (CR) mechanism to leverage reference alignments from OM for OV. This method (1) overcomes several pitfalls in using OM for OV tasks, (2) significantly reduces the number of matching candidates, and (3) improves overall OV performance. We showcase the use of our OM4OV framework in tracking version changes in a widely used ontology for the building industry.

We observe three opportunities for future work. (1) While OM has a universal evaluation measure, our proposed measures for OV have four sub-measures. We plan to investigate a merged formula for OV sub-measures in the future, for example, using a harmonic mean to combine sub-measures. (2) We limit the semantic analysis to noting changes in 1-hop subgraph relations that can be either ontology keywords (such as domain, range, and subclass) or domain concepts as object properties. A more complete analysis of structural evolution could employ description-logic-based reasoning for subsumption checking, consistency validation, and detection of semantically equivalent expressions with only syntactic differences. This kind of analysis has not been previously explored in OM and is not addressed in this study, but could be worthwhile further work. (3) We plan to apply the OM4OV framework in emerging domain ontologies that demonstrate regular changes over time. We are working with the Brick Schema development team to encourage adoption of our OM4OV framework.

\section*{Statement on Supplementary Materials and Additional Resources}
\phantomsection
\label{sec:supp-materials}

The following supplementary materials belong to this paper:\smallskip
\begin{enumerate}[(S1)]
\item
\emph{Software (OM4OV and Agent-OV):} {\small \url{https://github.com/qzc438/ontology-versioning}}\\
This artifact contains source code for OM4OV and Agent-OV, as well as the versioning datasets mentioned
in Section~\ref{sub-sec: dataset-construction}.
\item\emph{Software (Brick Schema Version Track):}
{\small \url{https://github.com/qzc438/brick-version-track}}\\
This artifact contains the source code for tracking Brick Schema~\cite{balaji2016brick} version changes mentioned in Section~\ref{sec: application}.
\end{enumerate}
\bigskip
The following resources were used in the mentioned parts of the paper:\smallskip
\begin{enumerate}[(R1)]
\item \emph{Software (Agent-OM)~\cite{github-R1}:} {\small \url{https://github.com/qzc438/ontology-llm}}\\
This artifact contains the source code for Agent-OM~\cite{qiang2023agent} mentioned in Sections~\ref{sub-sec: evaluation-system} and~\ref{sub-sec: results}.
\item \emph{Dataset (OAEI Datasets)~\cite{github-R2}:} {\small \url{https://dwslab.github.io/melt/track-repository}}\\
This artifact shows the Matching EvaLuation Toolkit (MELT)~\cite{hertling2019melt} repository to download
the original OAEI datasets for (S1) (retrieved April 1, 2025). According to the OAEI data policy, ``OAEI results and datasets, are publicly available, but subject to a use policy similar to \href{https://trec.nist.gov/results.html}{the one defined by NIST for TREC}. These rules apply to anyone using these data.'' Please find more details from the official website: \url{https://oaei.ontologymatching.org/doc/oaei-deontology.2.html} (retrieved April 1, 2025).
\item \emph{Dataset (Brick Schema Change Logs)~\cite{github-R3}:} {\small \url{https://github.com/BrickSchema/Brick/releases}}\\
This artifact shows the link to the Brick Schema~\cite{balaji2016brick} official repository to download the change logs for (S2) (retrieved April 1, 2025).
\end{enumerate}

\bibliography{qiang-bibliography-tgdk}

\end{document}